\documentclass[twoside]{article}

\usepackage[accepted]{aistats2024}
\usepackage[round]{natbib}

\usepackage[utf8]{inputenc} 
\usepackage[T1]{fontenc}    
\usepackage{hyperref}       
\usepackage{url}            
\usepackage{booktabs}       
\usepackage{amsfonts}       
\usepackage{nicefrac}       
\usepackage{microtype}      
\usepackage{xcolor}         

\usepackage{amssymb}
\usepackage{amsmath}
\usepackage{amsthm}
\usepackage{graphicx}
\usepackage{algorithm,algorithmic}
\usepackage{blkarray}
\usepackage{multirow}

\newtheorem{thm}{Theorem}
\newtheorem{lem}[thm]{Lemma}
\newtheorem{prop}[thm]{Proposition}

\newtheorem{defn}[thm]{Definition}

\newtheorem{exmp}{Example}

\newtheorem*{thm*}{Proposition}
\newtheorem*{prop*}{Proposition}
\newtheorem*{lem*}{Lemma}

\newcommand{\Sec}[1]{Section~\ref{#1}}

\definecolor{Gray}{gray}{0.8}

\renewcommand{\hat}{\widehat}
\renewcommand{\tilde}{\widetilde}
\renewcommand{\>}{{\rightarrow}}

\DeclareMathOperator*{\argmin}{argmin}

\newcommand{\R}{{\mathbb R}}

\newcommand{\N}{{\mathbb N}}
\renewcommand{\P}{{\mathbf P}}
\newcommand{\E}{{\mathbf E}}

\newcommand{\1}{{\mathbf 1}}

\newcommand{\A}{{\mathbf A}}
\newcommand{\cA}{{\mathcal A}}
\newcommand{\B}{{\mathcal B}}
\newcommand{\cC}{{\mathcal C}}
\newcommand{\C}{{\mathbf C}}

\renewcommand{\H}{{\mathcal H}}

\renewcommand{\L}{{\mathbf L}}

\renewcommand{\B}{{\mathbf B}}

\newcommand{\T}{{\mathbf T}}

\newcommand{\X}{{\mathcal X}}

\newcommand{\Y}{{\mathcal Y}}

\renewcommand{\a}{{\mathbf a}}

\newcommand{\p}{{\mathbf p}}

\newcommand{\w}{{\mathbf w}}
\newcommand{\x}{{\mathbf x}}

\newcommand{\zo}{\textup{\textrm{0-1}~}}

\newcommand{\regret}{\textup{\textrm{regret}}}

\newcommand{\ve}{\textup{\textrm{vec}}}
\newcommand{\bell}{{\boldsymbol \ell}}

\newcommand{\seta}{{\boldsymbol \eta}}
\newcommand{\bGamma}{{\boldsymbol \Gamma}}

\newcommand{\la}{\langle}
\newcommand{\ra}{\rangle}
\newcommand{\hattilde}[1]{\hat{\tilde{#1}}}

\makeatletter
  \g@addto@macro \normalsize{%
    \setlength\abovedisplayskip{4pt plus 0pt minus 0pt}%
    \setlength\belowdisplayskip{4pt plus 0pt minus 0pt}}%
\makeatother

\begin{document}

%
\runningtitle{Multiclass Learning from Noisy Labels for Non-decomposable Performance Measures}

%

\twocolumn[

\aistatstitle{Multiclass Learning from Noisy Labels for \\ Non-decomposable Performance Measures}

\aistatsauthor{Mingyuan Zhang and Shivani Agarwal}

\aistatsaddress{University of Pennsylvania\\
  Philadelphia, PA 19104 \\
  \texttt{\{myz, ashivani\}@seas.upenn.edu}} ]

\begin{abstract}
There has been much interest in recent years in learning good classifiers from data with noisy labels.
Most work on learning from noisy labels has focused on standard loss-based performance measures.
However, many machine learning problems require using \emph{non-decomposable} performance measures which cannot be expressed as the expectation or sum of a loss on individual examples; these include for example the H-mean, Q-mean and G-mean in class imbalance settings, and the Micro $F_1$ in information retrieval.
In this paper, we design algorithms to learn from noisy labels for two broad classes of multiclass non-decomposable performance measures, namely, monotonic convex and ratio-of-linear, which encompass all the above examples.
Our work builds on the Frank-Wolfe and Bisection based methods of Narasimhan et al. (2015).
In both cases, we develop noise-corrected versions of the algorithms under the widely studied family of class-conditional noise models.
We provide regret (excess risk) bounds for our algorithms, establishing that even though they are trained on noisy data, they are Bayes consistent in the sense that their performance converges to the optimal performance w.r.t. the clean (non-noisy) distribution.
Our experiments demonstrate the effectiveness of our algorithms in handling label noise.
\end{abstract}

\vspace{-12pt}
\section{\MakeUppercase{Introduction}}
\label{sec:intro}
\vspace{-6pt}

In many machine learning problems, the labels provided with the training data may be noisy. This can happen due to a variety of reasons, such as sensor measurement errors, human labeling errors, and data collection errors among others. Therefore, there has been much interest in recent years in learning good classifiers from data with noisy labels \citep{FrenayV14, Song2020,HanYLNTKS2020}.
Most work has focused on learning from noisy labels for standard loss-based performance measures; these include both the \zo loss and more general cost-sensitive losses, all of which are linear functions of the confusion matrix of a classifier.
However, many machine learning problems require using \emph{non-decomposable} performance measures which cannot be expressed as the expectation or sum of a loss on individual examples; these are general nonlinear functions of the confusion matrix, and include for example the H-mean, Q-mean and G-mean in class imbalance settings \citep{SunKW06,KennedyND09,LawrenceBBTG12,WangY12}, and the Micro $F_1$ in information retrieval \citep{ManningRS2008,KimWY13}.
In this paper, we design algorithms to learn from noisy labels for two broad classes of multiclass non-decomposable performance measures, namely, monotonic convex and ratio-of-linear, which encompass all the above examples.

\begin{table*}[!t]
\vspace{-16pt}
\begin{center}
\caption{
Position of Our Work Relative to Previous Work on Consistent Learning Under CCN Model
}
\label{tab:state}
\vspace{1pt}
\scalebox{1}{
\begin{tabular}{p{4cm}p{5.5cm}p{6cm}}
\hline
\textbf{Performance Measures} & \textbf{Standard (Non-noisy) Setting} & \textbf{Noisy Setting Under CCN Model} \\
\hline
Loss-based (linear) & Many algorithms including surrogate risk minimization algorithms & Many noise-corrected algorithms \citep{NatarajanDRT13,RooyenW17,PatriniRMNQ17,ZhangLA21} \\
Monotonic convex & Frank-Wolfe based method & \textcolor{red}{This work} \\
 & \citep{NarasimhanRS015} & \\
Ratio-of-linear & Bisection based method & \textcolor{red}{This work} \\
 & \citep{NarasimhanRS015} & \\
\hline
\end{tabular}
}
\end{center}
\vspace{-20pt}
\end{table*}

The main challenge in learning from noisy labels is to design algorithms which, given training data with noisy labels, can still learn accurate classifiers w.r.t. the clean/true distribution for a given target performance measure.
For loss-based (linear) performance measures, previous works have designed Bayes consistent algorithms so that, when given sufficient noisy training data, their performance converges to the Bayes optimal performance w.r.t. the clean distribution \citep{NatarajanDRT13,ScottBH13,Scott15,MenonROW15,LiuT16,PatriniNNC16,GhoshKS17,RooyenW17,PatriniRMNQ17,NatarajanDRT17,WangLT18,LiuG20,ZhangLA21,LiL00S21}.
In this work, we provide similarly Bayes consistent noise-corrected algorithms for multiclass monotonic convex and ratio-of-linear performance measures, under the widely studied family of class-conditional noise (CCN) models.
Our work builds on the Frank-Wolfe and Bisection based methods of \citet{NarasimhanRS015}, which were proposed for the standard (non-noisy) setting.
Table \ref{tab:state} summarizes the position of our work relative to other consistent algorithms under the CCN model.

Our key contributions include the following:
\vspace{-6pt}
\begin{itemize}
    \item \textbf{Algorithms:} We develop noise-corrected versions of the Frank-Wolfe and Bisection based algorithms for the families of monotonic convex and ratio-of-linear performance measures, respectively.
    \vspace{-4pt}
    \item \textbf{Theory:} While the noise corrections we introduce are fairly intuitive, establishing the correctness of the resulting algorithms is not trivial. We provide regret (excess risk) bounds for our algorithms, establishing that even though they are trained on noisy data, they are Bayes consistent in the sense that their performance converges to the optimal performance w.r.t. the clean (non-noisy) distribution. The bounds quantify the effect of label noise on the sample complexity. We also provide extended regret bounds that quantify the effect of using an estimated noise matrix.
    \vspace{-4pt}
    \item \textbf{Empirical validations:} We provide results of experiments on synthetic data verifying the sample complexity behavior of our algorithms, and also on real data comparing with previous baselines.
\end{itemize}

\vspace{-12pt}
\subsection{Related Work}
\vspace{-4pt}

\textbf{Consistent algorithms for binary/multiclass classification for non-decomposable performance measures in the standard (non-noisy) setting.}
Most work in this category has focused on binary classification, for a variety of performance measures, including F-measure \citep{YeCLC2012}, the arithmetic mean of the true positive and true negative rates (AM) \citep{MenonNAC13}, ratio-of-linear performance measures \citep{KoyejoNRD14,BaoS20}, and monotonic performance measures \citep{NarasimhanV014}. \cite{DembczynskiKKN17} revisited consistency analysis in binary classification for non-decomposable performance measures for two distinct settings and notions of consistency (Population Utility and Expected Test Utility).
For multiclass classification, \cite{NarasimhanRS015} developed a general framework for designing provably consistent algorithms for monotonic convex and ratio-of-linear performance measures; an extended version of this work also studies such performance measures in constrained learning settings \citep{NarasimhanRTKNA2022}.
\cite{ParambathUG14,KoyejoNRD15,NatarajanKRD16} also designed algorithms for some multiclass non-decomposable performance measures.
All of these works designed algorithms for standard (non-noisy) settings.
Our methods, which build on \citet{NarasimhanRS015}, are designed to correct for noisy labels for monotonic convex and ratio-of-linear performance measures, with provable consistency guarantees.

\textbf{Consistent algorithms for binary/multiclass learning from noisy labels for the \zo or cost-sensitive losses.}
For the CCN model in binary classification, many consistent algorithms have been proposed and analyzed \citep{NatarajanDRT13,ScottBH13,MenonROW15,LiuT16,PatriniNNC16,LiuG20}.
\cite{ScottBH13,Scott15,MenonROW15,LiuT16} also proposed consistent estimators for noise rates when they are not known (additional assumptions required).
\cite{ScottBH13,MenonROW15} studied the more general mutually contaminated distributions (MCD) noise model for binary classification, and proposed consistent algorithms.
\cite{NatarajanDRT17} studied cost-sensitive loss functions.
Progress has also been made in instance-dependent and label-dependent noise (ILN) model \citep{MenonRN18,ChengLRT20}.
For the multiclass CCN model, \cite{GhoshKS17,RooyenW17,PatriniRMNQ17,WangLT18,ZhangLA21,LiL00S21} proposed consistent algorithms.
All the methods above are designed to handle noisy labels for loss-based performance measures; our work, on the other hand, focuses on non-decomposable performance measures.

\textbf{Consistent algorithms for binary learning from noisy labels for non-decomposable performance measures.}
The method in \citet{ScottBH13} focused on the minmax error. \cite{MenonROW15} focused mostly on the balanced error (BER) and area under the ROC curve (AUC) metrics. Both studied the MCD model (which includes CCN model).
All these results are for binary classification. Our proposed algorithms, under the CCN model, are designed for monotonic convex and ratio-of-linear performance measures in both binary and multiclass classification settings.

\textbf{Performance measures in multi-label classification and structured prediction.}
There is also a line of work studying performance measures (e.g., Hamming loss and $F$-measure) in multi-label classification and structured prediction problems \citep{ZhangZ14,LiAWZB16,WangRZ17,ZhangRA20}, but those are distinct from (albeit related to) non-decomposable performance measures in multiclass classification settings as considered in this work.

\vspace{-8pt}
\subsection{Organization and Notation}
\vspace{-4pt}

\textbf{Organization.}
After preliminaries and background in \Sec{sec:prelim}, we
describe our noise-corrected algorithms for two broad classes of non-decomposable performance measures (monotonic convex and ratio-of-linear) in Section \ref{sec:monotonic} and Section \ref{sec:ratio-of-linear}, respectively.
Section \ref{sec:cons} provides consistency guarantees for our algorithms in the form of regret bounds.
Section \ref{sec:expts} summarizes our experiments.
Section \ref{sec:concl} concludes the paper.
All proofs can be found in Appendix \ref{appendix:section-proofs}.

\textbf{Notation.}
For an integer $n$, we denote by $[n]$ the set of integers $\{1,\ldots,n\}$, and by $\Delta_n$ 
the probability simplex $\{\p\in\R^n_+:\sum_{y=1}^n p_y=1\}$.
For a vector $\a$, we denote by $\|\a\|_p$ the $p$-norm of $\a$, and by $a_{j}$ the $j$-th entry of $\a$.
For a matrix $\A$, we denote by 
$\|\A\|_p$ the induced matrix $p$-norm of $\A$, and by
$\a_j$ the $j$-th column vector of $\A$. We use $A_{i,j}$ to denote the $(i,j)$-th entry of $\A$.
In addition, we use $\| \A \|_{\ve, p} = (\sum_{i,j} |
A_{i,j}^p|)^{1/p}$ for the matrix analogue of the vector $p$-norm.\footnote{Note that $\|\A\|_1$ and $\|
\A\|_\infty$ in \cite{NarasimhanRS015} are $\|\A\|_{\ve,1}$ and $\|\A\|_{\ve,\infty}$ in our notations. We choose to follow conventional definitions of the matrix norm in the literature instead.}
For matrices $\A, \B \in \R^{n \times n}$, we define $\la \A, \B \ra = \sum_{i,j} A_{i,j} B_{i,j}$.
The indicator function is $\1(\cdot)$.

\vspace{-8pt}
\section{\MakeUppercase{Preliminaries and Background}}
\label{sec:prelim}
\vspace{-6pt}

\textbf{Multiclass learning from noisy labels.}
Let $\X$ be the instance space, and $\Y$ be the label space. Without loss of generality, we assume $\Y = [n] = \{1,...,n\}$. There is an unknown distribution $D$ over $\X \times \Y$. In a standard multiclass learning problem, the learner is given labeled examples $(X,Y)$ drawn from $D$. However, when learning from noisy labels, the learner is only given noisy examples $(X, \tilde{Y})$, where $\tilde{Y}$ is the corresponding noisy label for $Y$. The learner's goal is to learn a classifier using the \emph{noisy} training sample, so that its performance is good w.r.t. the \emph{clean} distribution.

We consider the \emph{class-conditional noise} (CCN) model \citep{NatarajanDRT13,RooyenW17,PatriniRMNQ17}, in which a label $Y=y$ is switched by the noise process to $\tilde{Y}=\tilde{y}$ with probability $\P(\tilde{Y} = \tilde{y} | Y = y)$ that only depends on $y$ (and not on $x$). This noise can be fully described by a column stochastic matrix.

\begin{defn}[Class-conditional noise matrix]
The class-conditional noise matrix, $\T \in [0,1]^{n \times n}$, is column stochastic with entries $T_{i,j} = \P(\tilde{Y} = i | Y = j)$.
\end{defn}
\vspace{-6pt}

We assume $\T$ is invertible.
In practice, $\T$ often needs to be estimated; several methods have been developed to estimate $\T$ from the noisy sample \citep{XiaLW00NS19, YaoL0GD0S20, LiL00S21}. Our algorithms and theoretical guarantees work with both known $\T$ and estimated $\hat{\T}$.

We can then view the noisy training examples as being drawn i.i.d. from a \emph{noisy} distribution $\tilde{D}$ on $\X \times \Y$. Specifically, to generate $(X,\tilde{Y})$, an example $(X,Y)$ is firstly drawn according to $D$, and then $Y$ is switched to $\tilde{Y}$ according to noise matrix $\T$.

\textbf{Non-decomposable performance measure.}
To measure the performance of a classifier $h: \X \to [n]$, or more generally, a \emph{randomized classifier} $h : \X \to \Delta_n$ (which for a given instance $x$, predicts a label $y$ according to the probability specified by $h(x)$), we consider performance measures that are general functions of confusion matrices.

\begin{defn}[Confusion matrix]
The confusion matrix of a (possibly randomized) classifier $h$ w.r.t. a distribution $D$, denoted by $\C^{D} [h]$, has entries $C^D_{i,j} [h] = \P_{(X,Y) \sim D, Y' \sim h(X)}(Y = i, Y'=j)$,
where $Y' \sim h(X)$ denotes a random draw of label from distribution $h(X)$ when $h$ is randomized.
\end{defn}

\begin{defn}[Performance measure]
For any function $\psi : \R^{n \times n} \to \R_+$, define the $\psi$-performance measure of $h$ w.r.t. $D$ as
\begin{align*}
    \Psi_D^\psi [h] = \psi(\C^D[h]) \, .
\end{align*}
We adopt the convention that \emph{lower} values of $\Psi$ correspond to \emph{better} performance.
\end{defn}
\vspace{-6pt}

The following shows this formulation of performance measure includes the common loss-based performance measures (e.g., the \zo loss and cost-sensitive losses).

\begin{exmp}[$\L$-performance measures]
Consider a multiclass loss matrix $\L \in \R^{n \times n}$, where $L_{y, \hat{y}}$ is the loss incurred for predicting $\hat{y}$ when the true class is $y$. Then for a deterministic classifier $h$,
\begin{align*}
    \Psi_D^\L [h] = \E_{(X,Y) \sim D} \Big[ L_{Y,h(X)} \Big] =\la \L, \C^D[h] \ra \, .
\end{align*}
\end{exmp}
\vspace{-6pt}

In fact, loss-based performance measures are linear functions of confusion matrices. For nonlinear $\psi$, $\psi$-performance measures are \emph{non-decomposable}, i.e., they cannot be expressed as the expected loss on a new example drawn from $D$.
Common examples of such non-decomposable performance measures include Micro $F_1$ in information retrieval \citep{ManningRS2008,KimWY13}, H-mean, Q-mean and G-mean in class imbalance settings \citep{KennedyND09,LawrenceBBTG12,SunKW06,WangY12}, and others.\footnote{See Table 1 of \cite{NarasimhanRS015}.}

\textbf{Learning goal.}
Given a noisy training sample $\tilde{S}$ drawn according to the noisy distribution $\tilde{D}$, the goal of the learner is to learn a (randomized) classifier $h : \X \to \Delta_n$ that performs well w.r.t. $D$ for a pre-specified $\psi$-performance measure. In particular, we want the performance of $h$ to converge (in probability) to \emph{Bayes optimal $\psi$-performance} as the training sample size increases. Below we define Bayes optimal $\psi$-performance as the optimal value over \emph{feasible confusion matrices}.

\begin{defn}[Feasible confusion matrices]
\emph{Feasible confusion matrices} w.r.t. $D$ are all possible confusion matrices achieved by randomized classifiers. Define $\cC_D$ as the set of feasible confusion matrices w.r.t. $D$ as
\begin{align*}
    \cC_D = \{\C^D[h] : h : \X \to \Delta_n\} \, .
\end{align*}
\end{defn}
\vspace{-6pt}

We note that $\cC_D$ is a convex set (\cite{NarasimhanRS015}).

\begin{defn}[Bayes optimal $\psi$-performance]
For any function $\psi : \R^{n \times n} \to \R_+$, define the Bayes optimal $\psi$-performance w.r.t. $D$ as
\begin{align*}
    \Psi_D^{\psi,*} = \inf_{h : \X \to \Delta_n} \Psi_D^\psi[h] &= \inf_{h : \X \to \Delta_n} \psi(\C^D[h]) = \inf_{\C \in \cC_D} \psi(\C) \, .
\end{align*}
\end{defn}

In the following sections, we focus on two broad classes of non-decomposable performance measures, namely \emph{monotonic convex}
and \emph{ratio-of-linear}. The former includes H-mean, Q-mean and G-mean, and the latter includes Micro $F_1$.

\vspace{-8pt}
\section{\MakeUppercase{Monotonic Convex Performance Measures}}
\label{sec:monotonic}
\vspace{-6pt}

Our work develops noise-corrected versions of the algorithms of \citet{NarasimhanRS015}.
Below, we describe two key operations on which the algorithms in \cite{NarasimhanRS015} are built; we then describe our noise-corrected algorithm for monotonic convex performance measures. We will show how we use the noise matrix $\T$ to correct the two operations to learn from noisy labels.
We note that the noise correction operations work with estimated $\hat{\T}$ as well.
We start with the definition and some examples of monotonic convex performance measures.

\begin{defn}[Monotonic convex performance measures]
A performance measure $\psi : \R^{n \times n} \to \R_+$ is \emph{monotonic convex} if for any confusion matrix $\C$, $\psi(\C)$ is convex in $\C$, and monotonically (strictly) decreasing in $C_{i,i}$ and non-decreasing in $C_{i,j}$ for $i\ne j$.
\end{defn}

\begin{exmp}[H-mean, Q-mean and G-mean, all in loss forms]
H-mean: $\psi(\C) = 1 - n \big(\sum_{i=1}^n \frac{\sum_{j=1}^n C_{i,j}}{C_{i,i}} \big)^{-1}$, Q-mean: $\psi(\C) =  \sqrt{\frac{1}{n} \sum_{i=1}^n \big(1 - \frac{C_{i,i}}{\sum_{j=1}^n C_{i,j}} \big)^2}$, and G-mean: $\psi(\C) = 1 - \big(\prod_{i=1}^n \frac{C_{i,i}}{\sum_{j=1}^n C_{i,j}} \big)^{\frac{1}{n}}$.
\end{exmp}
\vspace{-6pt}

Next, we sketch the idea behind the algorithms in \cite{NarasimhanRS015}, and show how to introduce noise corrections to learn from noisy labels.
We first define the \emph{class probability function}.

\begin{defn}[Class probability function, class probability for short]
For $D$, the class probability function $\seta : \X \to \Delta_n$ is defined as $\eta_y(X) = \P(Y = y | X)$ for $y \in [n]$. Similarly for $\tilde{D}$, we define $\tilde{\seta} : \X \to \Delta_n$ as $\tilde{\eta}_{\tilde{y}}(X) = \P(\tilde{Y} = \tilde{y} | X)$ for $\tilde{y} \in [n]$.
\end{defn}

\textbf{Idea behind algorithms in the standard (non-noisy) setting \citep{NarasimhanRS015}.}
The algorithmic framework optimizes the non-decomposable performance measure $\psi$ of interest through an iterative approach (based on the Frank-Wolfe method for the monotonic convex case, and based on the bisection method for the ratio-of-linear case; details later), which in each iteration $t$, approximates the target performance measure $\psi$ by a linear loss-based performance measure $\L^t$.
Each iteration involves two key operations: OP1 and OP2.
\textbf{OP1} involves finding an optimal classifier for the current linear approximation $\L^t$.
This is done by using a class probability estimator (CPE) $\hat{\seta}$ learned from the (clean) training sample, and then defining classifier $\hat{g}^t : \X \to [n]$ as $\hat{g}^t(x) = \argmin_{y \in [n]} \hat{\seta}(x)^\top \bell_{y}^t$.
\textbf{OP2} involves estimating $\C^D[\hat{g}^t]$, the confusion matrix of $\hat{g}^t$ w.r.t. $D$, by $\hat{\C}^S[\hat{g}^t]$, the empirical confusion matrix of $\hat{g}^t$ w.r.t. sample $S = ((x_i,y_i))_{i=1}^m \sim D^m$ defined below:
\begin{align}
    \hat{\C}^S_{j,k}[\hat{g}^t] = \frac{1}{m} \sum_{i=1}^m \1 (y_i = j, \hat{g}^t(x_i) = k) \, . \label{eqn:empirical-conf}
\end{align}
\vspace{-6pt}

Note that $\hat{\C}^S[\hat{g}^t]$ converges to $\C^D[\hat{g}^t]$ as $m$ increases. (More specifically, to facilitate consistency analysis, the iterative algorithms split the training sample $S$ into $S_1$ and $S_2$. $S_1$ is used to learn a CPE model $\hat{\seta}$, and in each iterative step $t$, $S_2$ is used to calculate $\hat{\C}^{S_2}[\hat{g}^t]$ via OP2.)

\textbf{Noise-corrected algorithm for monotonic convex performance measures.}
We are now ready to describe our approach.
In learning from noisy labels, the algorithm only sees noisy sample $\tilde{S}$. Our approach is to introduce noise corrections to both OP1 and OP2, so the modified algorithm can still output a good classifier w.r.t. the clean distribution $D$.

\textbf{Noise-corrected OP1.} Recall OP1 involves finding an optimal classifier for a loss-based performance measure $\L^t$ w.r.t. $D$. To do so with a noisy sample, we propose to find an optimal classifier for a noise-corrected loss-based performance measure $(\L^t)' = (\T^\top)^{-1} \L^t$ w.r.t. $\tilde{D}$ according to the following proposition.

\begin{prop}
\label{prop:noisy-bayes-optimal-L}
Let  $\L' = (\T^\top)^{-1} \L$. Then any Bayes optimal classifier for $\L'$-performance w.r.t. $\tilde{D}$ is also Bayes optimal for $\L$-performance w.r.t. $D$.
\end{prop}
\vspace{-6pt}

This idea has also been used in multiclass noisy label settings with $\L$-performance \citep{RooyenW17,ZhangLA21}.

\textbf{Noise-corrected OP2.} Recall OP2 is to estimate $\C^D[\hat{g}^t]$. We need to do so with noisy sample $\tilde{S}$. We first observe a relation between clean confusion matrix $\C^D[\hat{g}^t]$ and noisy confusion matrix $\C^{\tilde{D}}[\hat{g}^t]$ under noise matrix $\T$.

\begin{prop}
\label{prop:relation-clean-noisy-conf}
For a given classifier $h$, the relation between \emph{clean} confusion matrix $\C^D$ and \emph{noisy} confusion matrix $\C^{\tilde{D}}$ under CCN matrix $\T$ is $\C^{\tilde{D}} [h] = \T \C^{D} [h]$.
\end{prop}
\vspace{-6pt}

So we propose to estimate $\C^D[\hat{g}^t]$ by $\T^{-1} \hat{\C}^{\tilde{S}}[\hat{g}^t]$. In Section \ref{sec:cons}, we will show this gives a consistent estimate, i.e., $\T^{-1} \hat{\C}^{\tilde{S}}[\hat{g}^t]$ converges to $\C^D[\hat{g}^t]$ as the size of $\tilde{S}$ increases.

We can now incorporate the noise-corrected OP1 and OP2 into the iterative algorithm based on Frank-Wolfe method \citep{FrankW1956,NarasimhanRS015}.
The noise-corrected algorithm is summarized in Algorithm \ref{alg:fw}.
This algorithm applies to monotonic convex performance measures $\psi$, such as H-mean, Q-mean and G-mean. It seeks to solve $\min_{\C \in \cC_D} \psi(\C)$ with the noisy sample $\tilde{S}$.
Note that the form $\nabla \psi(\cdot)$ in Line 7 comes from the form of Bayes optimal classifier for monotonic convex performance measures in the standard (non-noisy) setting (Theorem 13 of \cite{NarasimhanRS015}).
Specifically, Algorithm \ref{alg:fw} maintains $\C^t$ implicitly via $h^t$. At each step $t$, it applies noise-corrected OP1 and OP2 to construct a loss matrix $(\L^t)'$ and solve a linear minimization problem, and to compute an empirical confusion matrix.
The final randomized classifier $h^T$ is a convex combination of all the classifiers $h^0, h^1, ..., h^{T-1}$.
In Section \ref{sec:cons}, we will formally prove the noise-corrected algorithm is consistent.

\vspace{-8pt}
\section{\MakeUppercase{Ratio-of-linear Performance Measures}}
\label{sec:ratio-of-linear}
\vspace{-6pt}

\begin{figure}[t]
\vspace{-16pt}
\begin{algorithm}[H]
\begin{algorithmic}[1]
\STATE \textbf{Input:} 1) Performance measure $\psi:[0,1]^{n \times n} \to \R_+$ that is convex over $\cC_D$; 2) Noisy training sample $\tilde{S} = ((x_i, \tilde{y}_i))_{i=1}^m \in (\X \times \Y)^m$; 3) Noise matrix $\T$ (or estimated noise matrix $\hat{\T}$)
\STATE \textbf{Parameter:} Number of iterative steps $T \in \N$
\STATE Split $\tilde{S}$ into $\tilde{S}_1$ and $\tilde{S}_2$, each with size $\frac{m}{2}$
\STATE Run a CPE learner on $\tilde{S}_1$: $\hattilde{\seta}$ = CPE($\tilde{S}_1$)
\STATE \textbf{Initialize:} $h^0 : \X \to \Delta_n$, \, $\C^0 = \hat{\C}^{\tilde{S}_2}[h^0]$
\FOR{$t=1$ to $T$} 
\STATE Calculate noise-corrected loss-based performance measure $(\L^t)' = (\T^\top)^{-1} \nabla \psi(\T^{-1} \C^{t-1})$
\STATE Obtain $\hat{g}^t = x \mapsto \argmin_{y \in [n]} \hat{\tilde{\seta}}(x)^\top (\bell_{y}^{t})'$ and update $h^t = (1-\frac{2}{t+1}) h^{t-1} + \frac{2}{t+1} \hat{g}^t$
\STATE Calculate $\bGamma^t = \hat{\C}^{\tilde{S}_2}[\hat{g}^t]$ and update $\C^{t} = (1-\frac{2}{t+1}) \C^{t-1} + \frac{2}{t+1} \bGamma^t$
\ENDFOR
\STATE \textbf{Output:} $h^T$
\end{algorithmic}
\caption{\textsc{Noise-Corrected Frank-Wolfe (NCFW) Based Algorithm for Monotonic Convex Performance Measures} (See Section \ref{sec:monotonic} for details.)}\label{alg:fw}
\end{algorithm}
\vspace{-24pt}
\end{figure}

We now move to the next family of non-decomposable performance measures, namely ratio-of-linear performance measures. We start with the definition and an example. Then we will show how to use the noise-corrected OP1 and OP2 described in Section \ref{sec:monotonic} to build an algorithm to learn from noisy labels for ratio-of-linear performance measures.  We will also provide another view of the algorithm from the perspective of correcting the performance measure $\psi$.

\begin{defn}[Ratio-of-linear performance measures]
A performance measure $\psi : \R^{n \times n} \to \R_+$ is \emph{ratio-of-linear} if there are $\A, \B \in \R^{n \times n}$ such that for any confusion matrix $\C$, $\la \B, \C \ra > 0$ and $\psi(\C) = \frac{\la \A, \C \ra}{\la \B, \C \ra}$.
\end{defn}

\begin{exmp}[Micro $F_1$ in loss form]
Micro $F_1$: $\psi(\C) = 1 - \frac{2 \sum_{i=2}^n C_{i,i}}{2 - \sum_{i=1}^n C_{1,i} - \sum_{i=1}^n C_{i,1}}$.
\end{exmp}

\textbf{Noise-corrected algorithm for ratio-of-linear performance measures.}
The iterative algorithm based on Bisection method \citep{BoydL2004,NarasimhanRS015} follows broadly a similar idea as described in Section \ref{sec:monotonic}, so we can use the same noise-corrected OP1 and OP2 to modify the algorithm. The noise-corrected algorithm is summarized in Algorithm \ref{alg:bs}.
This algorithm applies to ratio-of-linear performance measures $\psi$, such as Micro $F_1$. It uses a binary search approach to find the minimum value of $\min_{\C \in \cC_D} \psi(\C)$.
Note that the form $\mathbf{A} - \gamma \mathbf{B}$ in Line 7 comes from the form of Bayes optimal classifier for ratio-of–linear performance measures in the standard (non-noisy) setting (Theorem 11 of \cite{NarasimhanRS015}).
Again, Algorithm \ref{alg:bs} maintains $\C^t$ implicitly via $h^t$. At each step $t$, it applies noise-corrected OP1 and OP2 to construct a loss matrix $(\L^t)'$ and solve a linear minimization problem, and to compute an empirical confusion matrix. The final classifier $h^T$ is deterministic.
In Section \ref{sec:cons}, we will formally prove the noise-corrected algorithm is consistent.

\begin{figure}[t]
\vspace{-16pt}
\begin{algorithm}[H]
\begin{algorithmic}[1]
\STATE \textbf{Input:} 1) Performance measure $\psi(\C) = \frac{\la \A, \C \ra}{\la \B, \C \ra}$ with $\A,\B \in \R^{n \times n}$; 2) Noisy training sample $\tilde{S} = ((x_i, \tilde{y}_i))_{i=1}^m \in (\X \times \Y)^m$; 3)
Noise matrix $\T$ (or estimated noise matrix $\hat{\T}$)
\STATE \textbf{Parameter:} Number of iterative steps $T \in \N$
\STATE Split $\tilde{S}$ into $\tilde{S}_1$ and $\tilde{S}_2$, each with size $\frac{m}{2}$
\STATE Run a CPE learner on $\tilde{S}_1$: $\hattilde{\seta}$ = CPE($\tilde{S}_1$)
\STATE \textbf{Initialize:} $h^0 : \X \to [n]$, \, $\alpha^0 = 0$, \, $\beta^0 = 1$
\FOR{$t=1$ to $T$}
\STATE Calculate noise-corrected loss $(\L^t)'= (\T^\top)^{-1} (\A - \gamma^t \B)$ where $\gamma^t = (\alpha^{t-1} + \beta^{t-1})/2$
\STATE Obtain $\hat{g}^t = x \mapsto \argmin_{y \in [n]} \hat{\tilde{\seta}}(x)^\top (\bell_{y}^{t})'$ and calculate $\bGamma^t = \hat{\C}^{\tilde{S}_2}[\hat{g}^t]$
\STATE \textbf{if} $\psi(\T^{-1} \bGamma^t) \le \gamma^t$ \textbf{then} $\alpha^t = \alpha^{t-1}, \beta^t = \gamma^t, h^t = \hat{g}^t$ \textbf{else} $\alpha^t = \gamma^t, \beta^t = \beta^{t-1}, h^t = h^{t-1}$
\ENDFOR
\STATE \textbf{Output:} $h^T$
\end{algorithmic}
\caption{\textsc{Noise-Corrected Bisection (NCBS) Based Algorithm for Ratio-of-linear Performance Measures} (See Section \ref{sec:ratio-of-linear} for details.)}\label{alg:bs}
\end{algorithm}
\vspace{-24pt}
\end{figure}

We also offer another view of Algorithm \ref{alg:bs} from the perspective of correcting $\psi$. In particular, we show that one can construct a noise-corrected performance measure $\tilde{\psi}$, which is also ratio-of-linear. Then one can simply optimize $\tilde{\psi}$ using a noisy sample to learn a classifier $h$, and the learned $h$ will also be optimal for the original performance measure $\psi$ w.r.t. the clean distribution $D$.
\begin{thm}[Form of Bayes optimal classifier for ratio-of-linear $\psi$ by correcting $\psi$]
\label{thm:noisy-bayes-ratio-of-linear-correcting-psi}
Consider ratio-of-linear performance measure $\psi(\C) = \frac{\la \A, \C \ra}{\la \B, \C \ra}$ with $\la \B, \C \ra > 0$ $\forall \C \in \cC_D$.
Define \emph{noise-corrected} performance measure $\tilde{\psi} : \R^{n \times n} \to \R_+$ by $\tilde{\psi} = \psi \circ \T^{-1}$.
Then $\tilde{\psi}(\tilde{\C}) = \frac{\la (\T^\top)^{-1} \A, \tilde{\C} \ra}{\la (\T^\top)^{-1} \B, \tilde{\C} \ra}$ with $\la (\T^\top)^{-1} \B, \tilde{\C} \ra > 0$ for all $\tilde{\C} \in \cC_{\tilde{D}}$. Moreover, any Bayes optimal classifier for $\tilde{\psi}$-performance w.r.t. $\tilde{D}$ is also Bayes optimal for $\psi$-performance w.r.t. $D$.
\end{thm}
\vspace{-6pt}

Therefore, one can view Algorithm \ref{alg:bs} as finding the Bayes optimal classifier for $\tilde{\psi}$ w.r.t. the noisy distribution $\tilde{D}$, which in turn is also Bayes optimal for $\psi$ w.r.t. the clean distribution $D$. This view is reminiscent of the Unbiased Estimator approach in \cite{RooyenW17} and Backward method in \cite{PatriniRMNQ17}, in which one optimizes noise-corrected surrogate losses using a noisy sample to learn classifiers that are optimal w.r.t. the clean distribution.

\vspace{-8pt}
\section{\MakeUppercase{Consistency and Regret Bounds}}
\label{sec:cons}
\vspace{-6pt}

In this section, we derive quantitative regret bounds for our noise-corrected algorithms. Our results show that when the CPE learner used in the algorithms is consistent (i.e., it converges to the noisy class probabilities), then the noise-corrected algorithms are consistent, i.e., they can output classifiers whose $\psi$-performance converges to the Bayes optimal $\psi$-performance w.r.t. $D$ as the size of the noisy training sample $\tilde{S}$ increases.
In addition, we provide regret bounds for our algorithms when estimated $\hat{\T}$ is used instead of $\T$.
To start, we formally define what it means for a learning algorithm to be $\psi$-consistent when learning from noisy labels. 

\begin{defn}[$\psi$-regret]
For any function $\psi : \R^{n \times n} \to \R_+$ and classifier $h : \X \to \Delta_n$, define $\psi$-regret of $h$ w.r.t. $D$ as the difference between $\psi$-performance of $h$ and the Bayes optimal $\psi$-performance:
$\regret_D^{\psi}[h] = \Psi_D^{\psi}[h] - \Psi_D^{\psi,*}$.
\end{defn}

\begin{defn}[$\psi$-consistent algorithm when learning from noisy labels]
For $\psi : \R^{n \times n} \to \R_+$, we say a multiclass algorithm $\cA : \cup_{m=1}^\infty \tilde{D}^m \to (\X \to \Delta_n)$, which given a noisy sample $\tilde{S}$ of size $m$ outputs a (randomized) classifier $\cA(\tilde{S})$, is \emph{consistent} for $\psi$ w.r.t. $D$ if for all $\epsilon > 0$:
\begin{align*}
    \P_{\tilde{S} \sim \tilde{D}^m} \big( \regret_D^{\psi}[\cA(\tilde{S})] > \epsilon \big) \to 0 \quad \text{as} \quad m \to \infty \, .
\end{align*}
\end{defn}

In Appendix \ref{appendix:section-proofs}, we provide guarantees for the noise-corrected OP1 and OP2 (Lemma \ref{lem:op1} and Lemma \ref{lem:op2}). They are used in deriving the following regret bounds.

\begin{thm}[$\psi$-regret bound for Algorithm \ref{alg:fw}]
\label{thm:fw-regret}
Let $\psi : \R^{n \times n} \to \R_+$ be monotonic convex over $\cC_D$, and $L$-Lipschitz and $\beta$-smooth w.r.t. $L_1$ norm.\footnote{A function $\psi$ is $\beta$-smooth if its gradient is $\beta$-Lipschitz.}
Noisy sample $\tilde{S} = ((x_i, \tilde{y}_i))_{i=1}^m \in (\X \times [n])^m$ is drawn randomly from $\tilde{D}^m$.
Let $\hattilde{\seta}:\X\>\Delta_n$ be the CPE model learned from $\tilde{S}_1$ as in Algorithm \ref{alg:fw}.
Then for $\delta \in (0,1]$, with probability at least $1-\delta$ (over $\tilde{S} \sim \tilde{D}^m$), we have
\begin{align*}
    &\regret_D^{\psi}[h^T] \le 4L \big\| \T^{-1} \big\|_1 \E_X \Big[ \big\| \hattilde{\seta}(X) - \tilde{\seta} (X) \big\|_1 \Big] + \frac{8 \beta}{T + 2} \\
    &~~~+ 4\sqrt{2}\beta n^3 C \big\| \T^{-1} \big\|_1 \sqrt{\frac{n^2 \log(n) \log(m) + \log(n^2/\delta)}{m}} \, ,
\end{align*}
where $C > 0$ is a distribution-independent constant.
\end{thm}

\begin{thm}[$\psi$-regret bound for Algorithm \ref{alg:bs}]
\label{thm:bs-regret}
Let $\psi(\C) = \frac{\la \A, \C \ra}{\la \B, \C \ra}$ for $\A, \B \in \R^{n \times n}$ with $\min_{\C \in \cC_D} \la \B, \C \ra \ge b$ for some $b > 0$.
Noisy sample $\tilde{S} = ((x_i, \tilde{y}_i))_{i=1}^m \in (\X \times [n])^m$ is drawn randomly from $\tilde{D}^m$.
Let $\hattilde{\seta}:\X\>\Delta_n$ be the CPE model learned from $\tilde{S}_1$ as in Algorithm \ref{alg:bs}.
Then for $\delta \in (0,1]$, with probability at least $1-\delta$ (over $\tilde{S} \sim \tilde{D}^m$), we have
\begin{align*}
    &\regret_D^{\psi}[h^T] \le 2 \tau \big\| \T^{-1} \big\|_1 \E_X \Big[ \big\| \hattilde{\seta}(X) - \tilde{\seta} (X) \big\|_1 \Big] + 2^{-T} \\
    &~~~+ 2\sqrt{2}\tau nC \big\| \T^{-1} \big\|_1 \sqrt{\frac{n^2 \log(n) \log(m) + \log(n^2/\delta)}{m}} \, ,
\end{align*}
where $\tau = \frac{1}{b}\big( \| \A \|_{\ve,1} + \| \B \|_{\ve,1} \big)$ and $C > 0$ is a distribution-independent constant.
\end{thm}
\vspace{-6pt}

In particular, using a strongly/strictly proper composite surrogate loss (e.g., multiclass logistic regression loss/cross entropy loss with softmax function) over a universal function class (with suitable regularization) to learn a CPE model ensures a consistent noisy class probability estimation, i.e., $\E_X \Big[ \big\| \hattilde{\seta}(X) - \tilde{\seta} (X) \big\|_1 \Big] \to 0$  as the sample size increases \citep{Agarwal14,WilliamsonVR16,ZhangLA21}. This leads to the convergence of the regret to zero as $m \to \infty, T \to \infty$.
Also, as the amount of label noise (captured by $\big\| \T^{-1} \big\|_1$) increases, the bounds get larger; one might therefore need a larger noisy sample size to achieve the same level of $\psi$-regret w.r.t. $D$.
Our synthetic experiments also confirm this sample complexity behavior.

\textbf{Regret bounds with estimated $\hat{\T}$.} When noise matrix $\T$ is not known, one may need to use estimated $\hat{\T}$. Several methods have been developed to estimate $\T$ from the noisy sample \citep{XiaLW00NS19, YaoL0GD0S20, LiL00S21}. Below, we provide regret bounds for our noise-corrected algorithms when estimated $\hat{\T}$ is used. They involve an additional factor $\big\| \hat{\T}^{-1} - \T^{-1} \big\|_1$ that quantifies the quality of the estimated $\hat{\T}$.

\begin{thm}[$\psi$-regret bound for Algorithm \ref{alg:fw} with estimated $\hat{\T}$]
\label{thm:fw-regret-add}
Let $\psi$, $\tilde{S}$ and $\hattilde{\seta}$ be specified as in Theorem \ref{thm:fw-regret}. Let $\hat{\T}$ be an estimate of $\T$. Then for $\delta \in (0,1]$, with probability at least $1-\delta$ (over $\tilde{S} \sim \tilde{D}^m$), we have
\begin{align*}
    &\regret_D^{\psi}[h^T] \le 4L \big\| \T^{-1} \big\|_1 \E_X \Big[ \big\| \hattilde{\seta}(X) - \tilde{\seta} (X) \big\|_1 \Big] + \frac{8 \beta}{T + 2} \\
    &~~~+ 4\sqrt{2}\beta n^3 C \big\| \T^{-1} \big\|_1 \sqrt{\frac{n^2 \log(n) \log(m) + \log(n^2/\delta)}{m}} \\
    &~~~+ (4L + 4\beta n^2 ) \big\| \hat{\T}^{-1} - \T^{-1} \big\|_1\, ,
\end{align*}
where $C > 0$ is a distribution-independent constant.
\end{thm}

\begin{thm}[$\psi$-regret bound for Algorithm \ref{alg:bs} with estimated $\hat{\T}$]
\label{thm:bs-regret-add}
Let $\psi$, $\tilde{S}$ and $\hattilde{\seta}$ be specified as in Theorem \ref{thm:bs-regret}. Let $\hat{\T}$ be an estimate of $\T$. Then for $\delta \in (0,1]$, with probability at least $1-\delta$ (over $\tilde{S} \sim \tilde{D}^m$), we have
\begin{align*}
    &\regret_D^{\psi}[h^T] \le 2 \tau \big\| \T^{-1} \big\|_1 \E_X \Big[ \big\| \hattilde{\seta}(X) - \tilde{\seta} (X) \big\|_1 \Big] + 2^{-T} \\
    &~~~+ 2\sqrt{2}\tau nC \big\| \T^{-1} \big\|_1 \sqrt{\frac{n^2 \log(n) \log(m) + \log(n^2/\delta)}{m}} \\
    &~~~+ 4\tau \big\| \hat{\T}^{-1} - \T^{-1} \big\|_1\, ,
\end{align*}
where $\tau = \frac{1}{b}\big( \| \A \|_{\ve,1} + \| \B \|_{\ve,1} \big)$ and $C > 0$ is a distribution-independent constant.
\end{thm}
\vspace{-6pt}

\vspace{-8pt}
\section{\MakeUppercase{Experiments}}
\label{sec:expts}
\vspace{-6pt}

We conducted two sets of experiments. In the first set of experiments, we generated synthetic data and tested the sample complexity behavior of our algorithms. In the second set of experiments, we used real data and compared our algorithms with other algorithms.
Our code is available at \url{https://github.com/moshimowang/noisy-labels-non-decomposable}.

\begin{figure}[t]
\vspace{-8pt}
\begin{center}
\scalebox{0.25}{\includegraphics{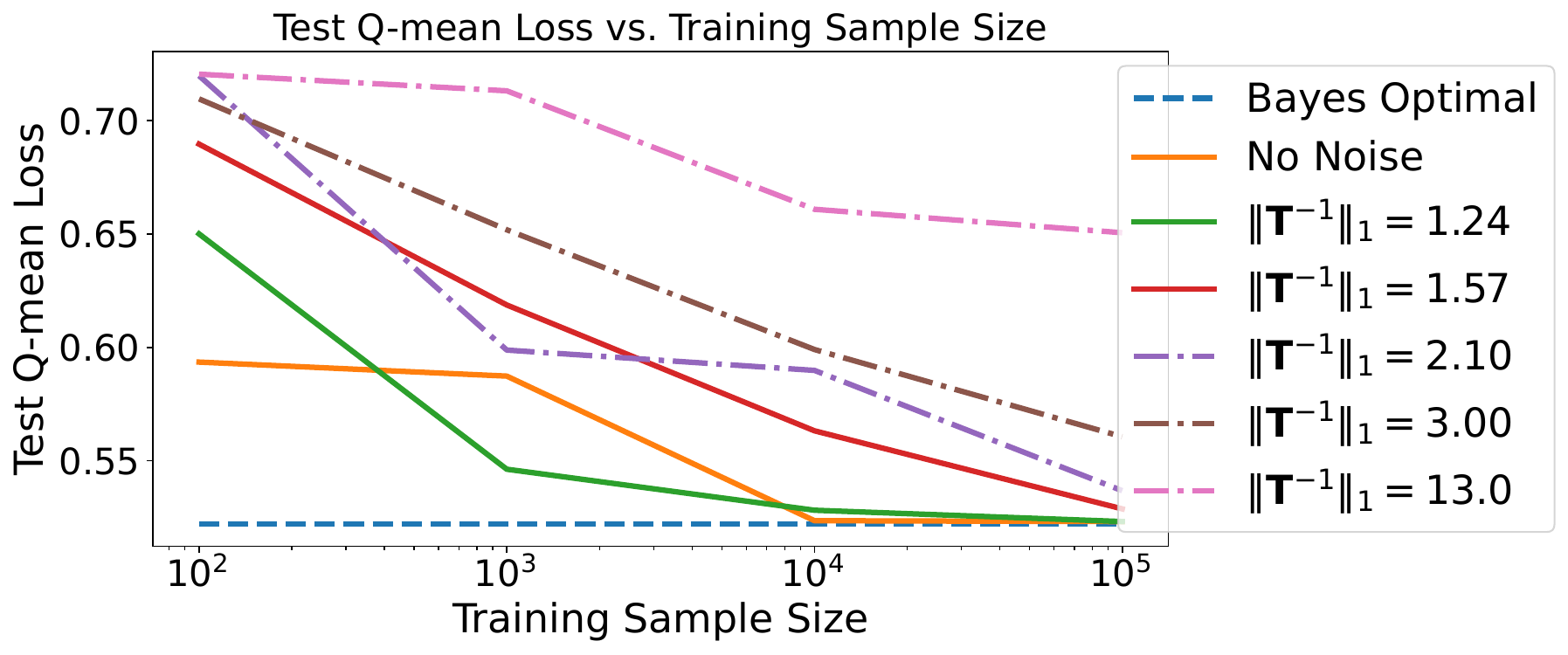}}\\
\scalebox{0.25}{\includegraphics{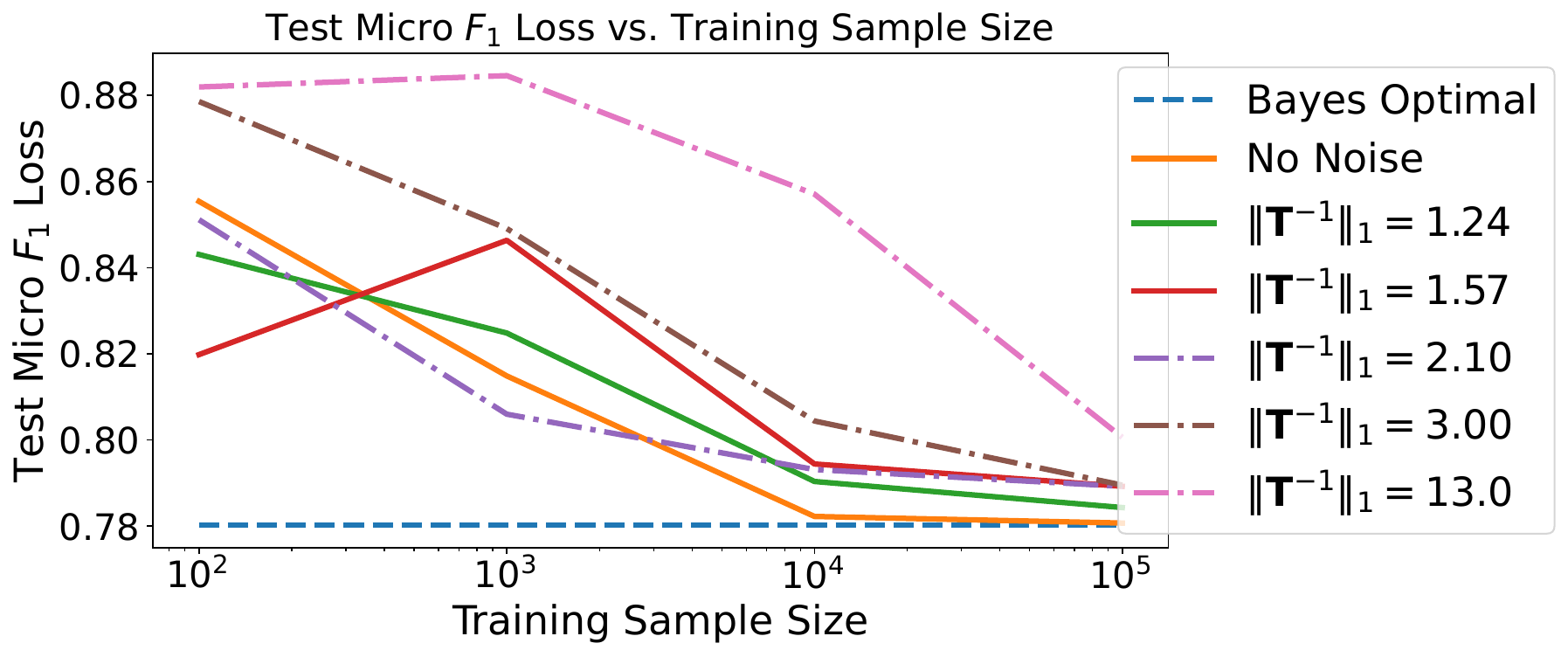}}
\vspace{-8pt}
\caption{
Sample Complexity Behavior of Our Noise-corrected Algorithms NCFW (top) and NCBS (bottom)
}
\label{fig:synthetic-multiclass}
\end{center}
\vspace{-20pt}
\end{figure}

\textbf{Sample complexity behavior.}
We tested the sample complexity behavior of our algorithm on synthetic data generated from a known distribution (see Appendix \ref{appendix:synthetic} for the data generating process).
We generated noise matrices by choosing a noise level $\sigma \in [0,1]$ and setting diagonal entries of $\T$ to $1-\sigma$ and off-diagonal entries of $\T$ to $\frac{\sigma}{2}$.
We tested the sample complexity behavior of our algorithms for a variety of noise matrices $\T$ with increasing values of  noise level $\sigma = 0.1, 0.2, 0.3, 0.4, 0.6$. The corresponding values of $\| \T^{-1} \|_1$ were also increasing.
The non-decomposable performance measures were Q-mean and Micro $F_1$. We applied Algorithm \ref{alg:fw} for Q-mean with $T=5000$ and Algorithm \ref{alg:bs} for Micro $F_1$ with $T=200$.
In both algorithms, the CPE learner was implemented by minimizing the multiclass logistic regression loss (aka. cross entropy loss with softmax function) over linear functions.
We ran the algorithms on noisy training samples with increasing sizes ($10^2, 10^3$, $10^4$, $10^5$), and measured the performance on a clean test set of $10^5$ examples.
The results are shown in Figure \ref{fig:synthetic-multiclass}.
The top plot shows results for Q-mean. The bottom plot shows results for Micro $F_1$.
We see that, as suggested by our regret bounds, as $\|\T^{-1}\|_1$ increases (i.e., more noise), the sample size required to achieve a given level of performance generally increases.

\begin{table*}[!t]
\vspace{-16pt}
\begin{center}
\caption{
Comparison with Other Algorithms for H-mean Loss
}
\label{tab:H-results}
\vspace{1pt}
\scalebox{0.9}{
\begin{tabular}{@{}llrrrr@{}}
\hline
\textbf{Data sets} & \textbf{Algorithms} & $\sigma=0.1$ & $\sigma=0.2$ & $\sigma=0.3$ & $\sigma=0.4$ \\
\hline
vehicle & FW & {0.255 (0.005)} & {0.267 (0.011)} & {0.309 (0.007)} & {0.373 (0.010)} \\
& NCFW & \textbf{0.254 (0.007)} & \textbf{0.266 (0.015)} & \textbf{0.307 (0.008)} & \textbf{0.338 (0.013)} \\
& NCLR-Backward & {0.482 (0.024)} & {0.573 (0.020)} & {0.512 (0.033)} & {0.508 (0.021)} \\
& NCLR-Forward & {0.512 (0.035)} & {0.563 (0.021)} & {0.570 (0.011)} & {0.563 (0.029)} \\
& NCLR-Plug-in & {0.515 (0.016)} & {0.567 (0.010)} & {0.517 (0.028)} & {0.540 (0.015)} \\
\hline
pageblocks & FW & {0.380 (0.041)} & {0.286 (0.011)} & {0.633 (0.097)} & {0.627 (0.066)} \\
& NCFW & \textbf{0.269 (0.017)} & \textbf{0.253 (0.006)} & \textbf{0.535 (0.019)} & \textbf{0.528 (0.034)} \\
& NCLR-Backward & {1.000 (0.000)} & {1.000 (0.000)} & {1.000 (0.000)} & {1.000 (0.000)} \\
& NCLR-Forward & {1.000 (0.000)} & {1.000 (0.000)} & {1.000 (0.000)} & {1.000 (0.000)} \\
& NCLR-Plug-in & {1.000 (0.000)} & {1.000 (0.000)} & {1.000 (0.000)} & {0.926 (0.066)} \\
\hline
satimage & FW & {0.188 (0.005)} & {0.224 (0.005)} & {0.247 (0.006)} & {0.340 (0.006)} \\
& NCFW & \textbf{0.186 (0.005)} & \textbf{0.222 (0.006)} & \textbf{0.230 (0.005)} & \textbf{0.300 (0.005)} \\
& NCLR-Backward & {0.556 (0.023)} & {0.630 (0.026)} & {0.685 (0.043)} & {0.960 (0.012)} \\
& NCLR-Forward & {0.542 (0.020)} & {0.522 (0.011)} & {0.612 (0.030)} & {0.877 (0.017)} \\
& NCLR-Plug-in & {0.679 (0.049)} & {0.793 (0.023)} & {0.854 (0.031)} & {0.902 (0.021)} \\
\hline
covtype & FW & {0.569 (0.001)} & {0.591 (0.001)} & {0.771 (0.010)} & {0.741 (0.009)} \\
& NCFW & \textbf{0.525 (0.001)} & \textbf{0.569 (0.001)} & \textbf{0.606 (0.002)} & \textbf{0.706 (0.004)} \\
& NCLR-Backward & {1.000 (0.000)} & {1.000 (0.000)} & {1.000 (0.000)} & {1.000 (0.000)} \\
& NCLR-Forward & {0.995 (0.001)} & {0.987 (0.002)} & {0.980 (0.002)} & {0.963 (0.005)} \\
& NCLR-Plug-in & {1.000 (0.000)} & {1.000 (0.000)} & {1.000 (0.000)} & {1.000 (0.000)} \\
\hline
abalone & FW & {0.806 (0.014)} & {0.799 (0.006)} & {0.812 (0.006)} & \textbf{0.801 (0.010)} \\
& NCFW & \textbf{0.797 (0.008)} & \textbf{0.795 (0.006)} & \textbf{0.804 (0.008)} & {0.814 (0.010)} \\
& NCLR-Backward & {1.000 (0.000)} & {1.000 (0.000)} & {1.000 (0.000)} & {1.000 (0.000)} \\
& NCLR-Forward & {1.000 (0.000)} & {1.000 (0.000)} & {1.000 (0.000)} & {1.000 (0.000)} \\
& NCLR-Plug-in & {1.000 (0.000)} & {1.000 (0.000)} & {1.000 (0.000)} & {1.000 (0.000)} \\
\hline
\end{tabular}
}
\end{center}
\vspace{-14pt}
\end{table*}

\begin{table*}[!t]
\vspace{-6pt}
\begin{center}
\caption{
Comparison with Other Algorithms for Micro $F_1$ Loss
}
\label{tab:F-results}
\vspace{1pt}
\scalebox{0.9}{
\begin{tabular}{@{}llrrrr@{}}
\hline
\textbf{Data sets} & \textbf{Algorithms} & $\sigma=0.1$ & $\sigma=0.2$ & $\sigma=0.3$ & $\sigma=0.4$ \\
\hline
vehicle & BS & {0.268 (0.007)} & {0.307 (0.006)} & {0.323 (0.013)} & {0.388 (0.012)} \\
& NCBS & \textbf{0.264 (0.008)} & \textbf{0.299 (0.006)} & \textbf{0.314 (0.011)} & \textbf{0.346 (0.009)} \\
& NCLR-Backward & {0.435 (0.020)} & {0.524 (0.006)} & {0.508 (0.027)} & {0.494 (0.028)} \\
& NCLR-Forward & {0.470 (0.031)} & {0.483 (0.020)} & {0.548 (0.016)} & {0.553 (0.030)} \\
& NCLR-Plug-in & {0.494 (0.022)} & {0.488 (0.023)} & {0.491 (0.025)} & {0.521 (0.015)} \\
\hline
pageblocks & BS & \textbf{0.231 (0.009)} & {0.323 (0.011)} & {0.862 (0.005)} & {0.899 (0.006)} \\
& NCBS & {0.251 (0.008)} & \textbf{0.261 (0.010)} & \textbf{0.320 (0.006)} & \textbf{0.404 (0.020)} \\
& NCLR-Backward & {0.515 (0.050)} & {0.457 (0.055)} & {0.756 (0.079)} & {0.510 (0.048)} \\
& NCLR-Forward & {0.823 (0.083)} & {0.880 (0.048)} & {0.743 (0.113)} & {0.832 (0.093)} \\
& NCLR-Plug-in & {0.609 (0.096)} & {0.595 (0.107)} & {0.568 (0.051)} & {0.795 (0.045)} \\
\hline
satimage & BS & {0.219 (0.004)} & {0.224 (0.002)} & {0.242 (0.002)} & {0.313 (0.003)} \\
& NCBS & {0.219 (0.004)} & {0.220 (0.003)} & {0.236 (0.002)} & {0.292 (0.002)} \\
& NCLR-Backward & \textbf{0.215 (0.004)} & {0.222 (0.003)} & {0.227 (0.004)} & {0.231 (0.003)} \\
& NCLR-Forward & {0.217 (0.003)} & \textbf{0.214 (0.002)} & \textbf{0.213 (0.003)} & \textbf{0.221 (0.003)} \\
& NCLR-Plug-in & {0.234 (0.002)} & {0.236 (0.003)} & {0.255 (0.002)} & {0.300 (0.003)} \\
\hline
covtype & BS & \textbf{0.361 (0.000)} & {0.355 (0.000)} & \textbf{0.362 (0.000)} & \textbf{0.362 (0.000)} \\
& NCBS & \textbf{0.361 (0.000)} & \textbf{0.352 (0.000)} & \textbf{0.362 (0.000)} & \textbf{0.362 (0.000)} \\
& NCLR-Backward & {0.384 (0.000)} & {0.385 (0.000)} & {0.388 (0.000)} & {0.390 (0.001)} \\
& NCLR-Forward & {0.384 (0.000)} & {0.380 (0.000)} & {0.381 (0.001)} & {0.382 (0.001)} \\
& NCLR-Plug-in & {0.398 (0.000)} & {0.396 (0.001)} & {0.397 (0.000)} & {0.397 (0.000)} \\
\hline
abalone & BS & {0.731 (0.007)} & {0.746 (0.005)} & {0.746 (0.003)} & \textbf{0.750 (0.002)} \\
& NCBS & \textbf{0.729 (0.007)} & \textbf{0.743 (0.005)} & \textbf{0.740 (0.003)} & {0.754 (0.001)} \\
& NCLR-Backward & {0.787 (0.005)} & {0.789 (0.007)} & {0.797 (0.010)} & {0.793 (0.011)} \\
& NCLR-Forward & {0.774 (0.007)} & {0.806 (0.004)} & {0.783 (0.003)} & {0.794 (0.009)} \\
& NCLR-Plug-in & {0.789 (0.005)} & {0.789 (0.009)} & {0.803 (0.007)} & {0.799 (0.010)} \\
\hline
\end{tabular}
}
\end{center}
\vspace{-20pt}
\end{table*}

\textbf{Comparison with other algorithms.}
We conducted experiments on several real data sets taken from UCI Machine Learning Repository \citep{uci}. Details of the data sets are in Appendix \ref{appendix:real}.
We compared our noise-corrected algorithms (NCFW and NCBS) with the baseline Frank-Wolfe (FW) and Bisection (BS) based methods of \citet{NarasimhanRS015,NarasimhanRTKNA2022} that were designed for the standard (non-noisy) learning setting, as well as various previously proposed noise-corrected versions of multiclass logistic regression (NCLR-Backward \citep{RooyenW17,PatriniRMNQ17}, NCLR-Forward \citep{PatriniRMNQ17}, and NCLR-Plug-in \citep{ZhangLA21}).
We used the authors' implementations for FW and BS.\footnote{\url{https://github.com/shivtavker/constrained-classification}.} To ensure a fair comparison, we also implemented our algorithms in the same framework.
Different variants of NCLR were implemented based on \cite{PatriniRMNQ17}.\footnote{\url{https://github.com/giorgiop/loss-correction}.}
A linear function class is used in all algorithms; see Appendix \ref{appendix:real} for more details.

To generate noise matrices $\T$, we chose a noise level $\sigma \in [0,1]$, set diagonal entries of $\T$ to $1-\sigma$, and set off-diagonal entries uniformly at random from $[0,1]$ so that each column of $\T$ sums to $1$. This makes sure that on average, $100 \sigma$ percent of clean labels were flipped to other labels, i.e., $\sigma \approx \frac{1}{m} \sum_{i=1}^m \1(y_i \ne \tilde{y}_i)$. Therefore, higher value of $\sigma$ means a higher noise level. We generated 4 noise matrices with $\sigma=0.1, 0.2, 0.3, 0.4$ according to this process. Training labels were flipped randomly according to the prescribed noise matrix $\T$.

We ran FW and NCFW for $T=5000$ iterative steps, and ran BS and NCBS for $T=200$ iterative steps. Performance of the learned model was then measured on a clean test set. The results are summarized in Table \ref{tab:H-results} (for H-mean loss) and Table \ref{tab:F-results} (for Micro $F_1$ loss), shown as the mean (with standard error of the mean in parentheses) over 5 random $7:3$ train-test splits. Higher $\sigma$ is a high noise level. For each data set and each noise level, the best performance is shown in bold font.
The results for G-mean loss and Q-mean loss can be found in Appendix \ref{appendix:real}.
As expected, in most cases, NCFW and NCBS outperform FW and BS, respectively, and they outperform variants of noise-corrected multiclass logistic regression as well.

\vspace{-8pt}
\section{\MakeUppercase{Conclusion}}
\label{sec:concl}
\vspace{-6pt}

We have provided the first known noise-corrected algorithms, NCFW and NCBS, for multiclass monotonic convex and ratio-of-linear performance measures under general class-conditional noise models.
We have also provided regret bounds for our algorithms showing that they are consistent w.r.t. the clean data distribution, and quantifying the effect of noise on their sample complexity.
Our experiments have demonstrated the effectiveness of our algorithms in handling label noise.
For settings where the noise matrix $\T$ may be unknown, approaches for estimating $\T$ have been proposed in the literature. These can be combined with our algorithms where needed, and we have also provided regret bounds for our algorithms when estimated $\hat{\T}$ is used.

\subsubsection*{Acknowledgements}
\vspace{-6pt}
The authors thank Harikrishna Narasimhan, Harish G. Ramaswamy, and Shiv Kumar Tavker for helpful discussions and for sharing the code of their work.
Thanks to the anonymous reviewers for several helpful comments and pointers. 
This material is based upon work supported in part by the US National Science Foundation (NSF) under Grant Nos.\ 1934876.
Any opinions, findings, and conclusions or recommendations expressed in this material are those of the authors and do not necessarily reflect the views of the National Science Foundation.

\newpage


\section*{Checklist}



 \begin{enumerate}

 \item For all models and algorithms presented, check if you include:
 \begin{enumerate}
   \item A clear description of the mathematical setting, assumptions, algorithm, and/or model. [Yes]
   \item An analysis of the properties and complexity (time, space, sample size) of any algorithm. [Yes]
   \item (Optional) Anonymized source code, with specification of all dependencies, including external libraries. [Yes]
 \end{enumerate}

 \item For any theoretical claim, check if you include:
 \begin{enumerate}
   \item Statements of the full set of assumptions of all theoretical results. [Yes]
   \item Complete proofs of all theoretical results. [Yes]
   \item Clear explanations of any assumptions. [Yes]     
 \end{enumerate}

 \item For all figures and tables that present empirical results, check if you include:
 \begin{enumerate}
   \item The code, data, and instructions needed to reproduce the main experimental results (either in the supplemental material or as a URL). [Yes]
   \item All the training details (e.g., data splits, hyperparameters, how they were chosen). [Yes]
         \item A clear definition of the specific measure or statistics and error bars (e.g., with respect to the random seed after running experiments multiple times). [Yes]
         \item A description of the computing infrastructure used. (e.g., type of GPUs, internal cluster, or cloud provider). [Yes; see Appendix]
 \end{enumerate}

 \item If you are using existing assets (e.g., code, data, models) or curating/releasing new assets, check if you include:
 \begin{enumerate}
   \item Citations of the creator If your work uses existing assets. [Yes]
   \item The license information of the assets, if applicable. [Not Applicable]
   \item New assets either in the supplemental material or as a URL, if applicable. [Not Applicable]
   \item Information about consent from data providers/curators. [Not Applicable]
   \item Discussion of sensible content if applicable, e.g., personally identifiable information or offensive content. [Not Applicable]
 \end{enumerate}

 \item If you used crowdsourcing or conducted research with human subjects, check if you include:
 \begin{enumerate}
   \item The full text of instructions given to participants and screenshots. [Not Applicable]
   \item Descriptions of potential participant risks, with links to Institutional Review Board (IRB) approvals if applicable. [Not Applicable]
   \item The estimated hourly wage paid to participants and the total amount spent on participant compensation. [Not Applicable]
 \end{enumerate}

 \end{enumerate}

\newpage
\appendix
\onecolumn

\aistatstitle{Multiclass Learning from Noisy Labels for \\ Non-decomposable Performance Measures \\
Supplementary Materials}

\section{Proofs}
\label{appendix:section-proofs}

\subsection{Proofs for Section \ref{sec:monotonic}}

\textbf{Proof of Proposition \ref{prop:noisy-bayes-optimal-L}}.

\begin{proof}

Let $h^*$ be a Bayes optimal classifier for $(\T^\top)^{-1} \L$-performance w.r.t. $\tilde{D}$. So
\begin{align*}
\inf_{\tilde{\C} \in \cC_{\tilde{D}}} \la (\T^\top)^{-1} \L, \tilde{\C} \ra &= \la (\T^\top)^{-1} \L, \C^{\tilde{D}}[h^*] \ra \\
&= \la \L, \T^{-1} \C^{\tilde{D}}[h^*] \ra \\
&= \la \L, \C^D[h^*] \ra \, ,
\end{align*}
where we have used properties of the adjoint in the second ``=''.

Note that for any $\C \in \cC_D$, we have
\begin{align*}
    \la \L, \C \ra &= \la (\T^\top)^{-1} \L, \T \C \ra \\
    &\ge \inf_{\tilde{\C} \in \cC_{\tilde{D}}} \la (\T^\top)^{-1} \L, \tilde{\C} \ra \\
    &= \la \L, \C^D[h^*] \ra \, .
\end{align*}
So $h^*$ is also Bayes optimal for $\L$-performance w.r.t. $D$, i.e., $\la \L, \C^D[h^*] \ra = \Psi_D^{\L,*}$.
\end{proof}

\textbf{Proof of Proposition \ref{prop:relation-clean-noisy-conf}.}

\begin{proof}
\begin{align*}
    C^{\tilde{D}}_{i,j}[h]
    &= \P(\tilde{Y} = i, h(X) = j) \\
    &= \P(\tilde{Y} = i | h(X) = j) \P(h(X) = j) \\
    &= \sum_{k \in [n]} \P(\tilde{Y} = i, Y = k | h(X) = j) \P(h(X) = j) \\
    &= \sum_{k \in [n]} \P(\tilde{Y} = i | Y = k) \P(Y = k | h(X) = j) \P(h(X) = j) \\
    &= \sum_{k \in [n]} T_{i,k} \cdot \P(Y = k | h(X) = j) \P(h(X) = j) \\
    &= \sum_{k \in [n]} T_{i,k} \cdot \P(Y = k, h(X) = j) \\
    &= \sum_{k \in [n]} T_{i,k} \cdot C^D_{kj} [h] \, .
\end{align*}
\end{proof}

\subsection{Proofs for Section \ref{sec:ratio-of-linear}}

\textbf{Proof of Theorem \ref{thm:noisy-bayes-ratio-of-linear-correcting-psi}.}

\begin{proof}
By property of adjoint, we have
\begin{align*}
    \tilde{\psi}(\tilde{\C}) = \psi \circ \T^{-1} (\tilde{\C}) = \psi\Big( \T^{-1} \tilde{\C} \Big) = \frac{\la \A, \T^{-1} \tilde{\C} \ra}{\la \B, \T^{-1} \tilde{\C} \ra} = \frac{\la (\T^\top)^{-1} \A, \tilde{\C} \ra}{\la (\T^\top)^{-1} \B, \tilde{\C} \ra} \, .
\end{align*}
For all $\tilde{\C} \in \cC_{\tilde{D}}$, there exists $\C \in \cC_D$ such that $\tilde{\C} = \T \C$. So,
\begin{align*}
    \la (\T^\top)^{-1} \B, \tilde{\C} \ra = \la \B, \C \ra > 0 \, .
\end{align*}
This shows $\la (\T^\top)^{-1} \B, \tilde{\C} \ra > 0$ for all $\tilde{\C} \in \cC_{\tilde{D}}$.

Let $h^*$ be a Bayes optimal classifier for $\tilde{\psi}$-performance w.r.t. $\tilde{D}$ (the existence of such a classifier is guaranteed by Theorem 11 of \cite{NarasimhanRS015}). So,
\begin{align*}
    \inf_{\tilde{C} \in \cC_{\tilde{D}}} \tilde{\psi}(\tilde{\C}) &= \tilde{\psi}(\C^{\tilde{D}}[h^*]) \\
    &= \frac{\la (\T^\top)^{-1} \A, \C^{\tilde{D}}[h^*] \ra}{\la (\T^\top)^{-1} \B, \C^{\tilde{D}}[h^*] \ra} \\
    &= \frac{\la (\T^\top)^{-1} \A, \T \C^{D}[h^*] \ra}{\la (\T^\top)^{-1} \B, \T \C^{D}[h^*] \ra} \\
    &= \frac{\la \A, \C^{D}[h^*] \ra}{\la \B, \C^{D}[h^*] \ra} \, .
\end{align*}
 For all $\C \in \cC_D$,
\begin{align*}
    \psi(\C) &= \frac{\la \A, \C \ra}{\la \B, \C \ra} \\
    &\ge \inf_{\C' \in \cC_{D}} \frac{\la \A, \C' \ra}{\la \B, \C' \ra} \\
    &= \inf_{\C' \in \cC_{D}} \frac{\la (\T^\top)^{-1} \A, \T \C' \ra}{\la (\T^\top)^{-1} \B, \T \C' \ra} \\
    &= \inf_{\tilde{\C} \in \cC_{\tilde{D}}} \tilde{\psi}(\tilde{\C}) \\
    &= \frac{\la \A, \C^{D}[h^*] \ra}{\la \B, \C^{D}[h^*] \ra} \, .
\end{align*}
This shows $h^*$ is also Bayes optimal for $\psi$-performance w.r.t. $D$.
\end{proof}

\subsection{Proofs for Section \ref{sec:cons}}

\begin{lem}[Guarantee for noise-corrected OP1]
\label{lem:op1}
Let $\hattilde{\seta}:\X\>\Delta_n$ be the CPE model learned from noisy sample $\tilde{S}$. Then
\begin{align*}
&\E_X \Big[ \big\| \T^{-1} \hattilde{\seta}(X) - \seta (X) \big\|_1 \Big] \\
&~~~\le \big\| \T^{-1} \big\|_1 \cdot \E_X \Big[ \big\| \hattilde{\seta}(X) - \tilde{\seta} (X) \big\|_1 \Big] \, .
\end{align*}
\end{lem}

\textbf{Proof of Lemma \ref{lem:op1}.}

\begin{proof}
\begin{align*}
   \E_X \Big[ \big\| \T^{-1} \hattilde{\seta}(X) - \seta (X) \big\|_1 \Big] = \E_X \Big[ \big\| \T^{-1} \hattilde{\seta}(X) - \T^{-1} \tilde{\seta} (X) \big\|_1 \Big] \le \big\| \T^{-1} \big\|_1 \cdot \E_X \Big[ \big\| \hattilde{\seta}(X) - \tilde{\seta} (X) \big\|_1 \Big] \, .
\end{align*}
\end{proof}

\begin{lem}[Guarantee for noise-corrected OP2]
\label{lem:op2}
For $h : \X \to \Delta_n$, let $\hat{\C}^{\tilde{S}}[h]$ be the empirical confusion matrix w.r.t. noisy sample $\tilde{S}$ (computed similarly as $\hat{\C}^{S}[h]$ in Eq. \eqref{eqn:empirical-conf}). Then
\begin{align*}
&\big\| \C^D[h] - \T^{-1} \hat{\C}^{\tilde{S}}[h] \big\|_{\ve,\infty} \\
&~~~\le n \big\| \T^{-1} \big\|_1 \cdot \big\| \C^{\tilde{D}}[h] - \hat{\C}^{\tilde{S}}[h] \big\|_{\ve,\infty} \, .
\end{align*}
\end{lem}

\textbf{Proof of Lemma \ref{lem:op2}.}

\begin{proof}
\begin{align*}
\big\| \C^D[h] - \T^{-1} \hat{\C}^{\tilde{S}}[h] \big\|_{\ve,\infty} &= \big\| \T^{-1} \C^{\tilde{D}}[h] - \T^{-1} \hat{\C}^{\tilde{S}}[h] \big\|_{\ve,\infty} \\
&\le \big\| \T^{-1} \C^{\tilde{D}}[h] - \T^{-1} \hat{\C}^{\tilde{S}}[h] \big\|_{1} \\
&\le \big\| \T^{-1} \big\|_1 \cdot \big\| \C^{\tilde{D}}[h] - \hat{\C}^{\tilde{S}}[h] \big\|_1 \\
&\le n \big\| \T^{-1} \big\|_1 \cdot \big\| \C^{\tilde{D}}[h] - \hat{\C}^{\tilde{S}}[h] \big\|_{\ve,\infty}\, .
\end{align*}
\end{proof}

\textbf{Notes for Lemma \ref{lem:op1} and Lemma \ref{lem:op2}:}
In Lemma \ref{lem:op1}, $\T^{-1} \hattilde{\seta}(x)$ might be viewed as an estimate for $\seta(x)$. If the CPE model used is consistent (i.e., $\E_X \Big[ \big\| \hattilde{\seta}(X) - \tilde{\seta} (X) \big\|_1 \Big] \to 0$ as the sample size increases), then this estimation is consistent as well.
Similarly, in Lemma \ref{lem:op2}, $\T^{-1} \hat{\C}^{\tilde{S}}[h]$ might be viewed as an estimate for $\C^D[h]$. Because the confusion matrix estimator in Eq. \eqref{eqn:empirical-conf} is consistent (i.e., $\big\| \C^{\tilde{D}}[h] - \hat{\C}^{\tilde{S}}[h] \big\|_{\ve,\infty} \to 0$  as the sample size increases) as shown in Lemma 15 of \cite{NarasimhanRS015}, this estimation is also consistent.
$\big\| \T^{-1} \big\|_1$ might be viewed as a constant capturing the overall amount of label noise.

\textbf{Notes for our proof of Theorem \ref{thm:fw-regret}:}
Our proof of Theorem \ref{thm:fw-regret} uses Lemma 14, Lemma 15 and Theorem 16 in \cite{NarasimhanRS015}, along with their proofs. We include a proposition below that summarizes the key aspects of Lemma 14, Lemma 15 and Theorem 16 in \cite{NarasimhanRS015} that we make use of (with slight modification in order for it to be consistent with our notations).

\begin{prop}[$\psi$-regret bound of Frank-Wolfe based algorithm in the non-noisy setting; Theorem 16 in \cite{NarasimhanRS015}]
\label{thm:fw-regret-standard}
Let $\psi : \R^{n \times n} \to \R_+$ be monotonic convex over $\cC_D$, and $L$-Lipschitz and $\beta$-smooth w.r.t. $L_1$ norm.
Let clean sample $S = ((x_i, y_i))_{i=1}^m \in (\X \times [n])^m$ be drawn randomly from $D^m$.
Let $\hat{\seta}:\X\>\Delta_n$ be the CPE model learned from $S_1$ as in the Frank-Wolfe based algorithm, and $h^{\mathrm{FW}}_S : \X \to \Delta_n$ be the classifier returned after $T$ iterations.
Let $\delta \in (0,1]$. Then with probability $\ge 1-\delta$ (over $S \sim D^m$),
\begin{align*}
    &\regret_D^{\psi}[h^{\mathrm{FW}}_S] \le 4L \E_X \Big[ \big\| \hat{\seta}(X) - \seta (X) \big\|_1 \Big] + \frac{8 \beta}{T + 2} \\
    &~~~+ 4\beta n^2 \sup_{h \in \H_{\hat{\seta}}} \big\| \C^D[h] -  \hat{\C}^{S_2}[h]\big\|_{\ve,\infty} \\
    &~~~\le 4L \E_X \Big[ \big\| \hat{\seta}(X) - \seta (X) \big\|_1 \Big] + \frac{8 \beta}{T + 2} \\
    &~~~+ 4\sqrt{2}\beta n^2 C \sqrt{\frac{n^2 \log(n) \log(m) + \log(n^2/\delta)}{m}} \, ,
\end{align*}
where $C > 0$ is a distribution-independent constant, and $\H_{\hat{\seta}} = \{ h : \X \to [n], h(x) = \argmin_{y \in [n]} \hat{\seta}(x)^\top \bell_{y}, \L \in \R^{n \times n} \}$.
The second `$\le$' was obtained by Lemma 15 of \cite{NarasimhanRS015} and $|S_2| = m/2$.
\end{prop}

\textbf{Proof of Theorem \ref{thm:fw-regret}.}

\begin{proof}
In Algorithm \ref{alg:fw}, we only have noisy sample $\tilde{S}$ that is split into $\tilde{S}_1$ and $\tilde{S}_2$.
We implicitly estimate $\seta$ by $\T^{-1} \circ \hattilde{\seta}$, where $\hattilde{\seta}:\X\>\Delta_n$ is a CPE model learned from $\tilde{S}_1$. Lemma \ref{lem:op1} shows an additional factor of $\big\| \T^{-1} \big\|_1$ as a price paid to learn from a noisy sample instead of a clean one.
For a classifier $h$, we estimate $\C^{D}[h]$ by $\T^{-1} \hat{\C}^{\tilde{S}_2}[h]$, where $\hat{\C}^{\tilde{S}_2}[h]$ is the empirical confusion matrix learned from $\tilde{S}_2$. Lemma \ref{lem:op2} shows an additional factor of $n \big\| \T^{-1} \big\|_1$ as a cost to learn from a noisy sample instead of a clean one.
Note that $\H_{\hat{\seta}} = \H_{\hattilde{\seta}}$ for $\hat{\seta} = \T^{-1} \circ \hattilde{\seta}$.
Chaining this reasoning with Proposition \ref{thm:fw-regret-standard} establishes the claim.
\end{proof}

\textbf{Notes for our proof of Theorem \ref{thm:bs-regret}:}
Our proof of Theorem \ref{thm:bs-regret} uses Lemma 14, Lemma 15 and Theorem 17 in \cite{NarasimhanRS015}, along with their proofs. We include a proposition below that summarizes the key aspects of Lemma 14, Lemma 15 and Theorem 17 in \cite{NarasimhanRS015} that we make use of (with slight modification in order for it to be consistent with our notations).

\begin{prop}[$\psi$-regret bound for bisection based algorithm in the non-noisy setting; Theorem 17 in \cite{NarasimhanRS015}]
\label{thm:bs-regret-standard}
Let $\psi(\C) = \frac{\la \A, \C \ra}{\la \B, \C \ra}$ for $\A, \B \in \R^{n \times n}$ with $\min_{\C \in \cC_D} \la \B, \C \ra \ge b$ for some $b > 0$.
Let clean sample $S = ((x_i, y_i))_{i=1}^m \in (\X \times [n])^m$ be drawn randomly from $D^m$.
Let $\hat{\seta}:\X\>\Delta_n$ be the CPE model learned from $S_1$ as in the bisection based algorithm, and $h^{\mathrm{BS}}_S : \X \to \Delta_n$ be the classifier returned after $T$ iterations.
Let $\delta \in (0,1]$. Then with probability $\ge 1-\delta$ (over $S \sim D^m$),
\begin{align*}
    &\regret_D^{\psi}[h^{\mathrm{BS}}_S] \le 2 \tau \E_X \Big[ \big\| \hat{\seta}(X) - \seta (X) \big\|_1 \Big] + 2^{-T} \\
    &~~~+ 2\tau \sup_{h \in \H_{\hat{\seta}}} \big\| \C^D[h] -  \hat{\C}^{S_2}[h]\big\|_{\ve,\infty} \\
    &~~~\le 2 \tau \E_X \Big[ \big\| \hat{\seta}(X) - \seta (X) \big\|_1 \Big] + 2^{-T} \\
    &~~~+ 2\sqrt{2} C \tau \sqrt{\frac{n^2 \log(n) \log(m) + \log(n^2/\delta)}{m}} \, ,
\end{align*}
where $\tau = \frac{1}{b}\big( \| \A \|_{\ve,1} + \| \B \|_{\ve,1} \big)$, $C > 0$ is a distribution-independent constant, and $\H_{\hat{\seta}} = \{ h : \X \to [n], h(x) = \argmin_{y \in [n]} \hat{\seta}(x)^\top \bell_{y}, \L \in \R^{n \times n} \}$.
The second `$\le$' was obtained by Lemma 15 of \cite{NarasimhanRS015} and $|S_2| = m/2$.
\end{prop}

\textbf{Proof of Theorem \ref{thm:bs-regret}.}

\begin{proof}
In Algorithm \ref{alg:bs}, we only have noisy sample $\tilde{S}$ that is split into $\tilde{S}_1$ and $\tilde{S}_2$.
We implicitly estimate $\seta$ by $\T^{-1} \circ \hattilde{\seta}$, where $\hattilde{\seta}:\X\>\Delta_n$ is a CPE model learned from $\tilde{S}_1$. Lemma \ref{lem:op1} shows an additional factor of $\big\| \T^{-1} \big\|_1$ as a price paid to learn from a noisy sample instead of a clean one.
For a classifier $h$, we estimate $\C^{D}[h]$ by $\T^{-1} \hat{\C}^{\tilde{S}_2}[h]$, where $\hat{\C}^{\tilde{S}_2}[h]$ is the empirical confusion matrix learned from $\tilde{S}_2$. Lemma \ref{lem:op2} shows an additional factor of $n \big\| \T^{-1} \big\|_1$ as a cost to learn from a noisy sample instead of a clean one.
Note that $\H_{\hat{\seta}} = \H_{\hattilde{\seta}}$ for $\hat{\seta} = \T^{-1} \circ \hattilde{\seta}$.
Chaining this reasoning with Proposition \ref{thm:bs-regret-standard} establishes the claim.
\end{proof}

\begin{lem}[Guarantee for noise-corrected OP1 with estimated $\hat{\T}$]
\label{lem:op1-add}
Let $\hattilde{\seta}:\X\>\Delta_n$ be the CPE model learned from noisy sample $\tilde{S}$. Let $\hat{\T}$ be an estimate of $\T$. Then
\begin{align*}
&\E_X \Big[ \big\| \hat{\T}^{-1} \hattilde{\seta}(X) - \seta (X) \big\|_1 \Big] \\
&\le \big\| \T^{-1} \big\|_1 \cdot \E_X \Big[ \big\| \hattilde{\seta}(X) - \tilde{\seta} (X) \big\|_1 \Big] + \big\| \hat{\T}^{-1} - \T^{-1} \big\|_1 \, .
\end{align*}
\end{lem}

\textbf{Proof of Lemma \ref{lem:op1-add}.}

\begin{proof}
\begin{align*}
   &\E_X \Big[ \big\| \hat{\T}^{-1} \hattilde{\seta}(X) - \seta (X) \big\|_1 \Big] \\
   &= \E_X \Big[ \big\| \hat{\T}^{-1} \hattilde{\seta}(X) - \T^{-1} \tilde{\seta} (X) \big\|_1 \Big] \\
   &= \E_X \Big[ \big\| \hat{\T}^{-1} \hattilde{\seta}(X) - \T^{-1} \tilde{\seta} (X) + \T^{-1} \hattilde{\seta}(X) - \T^{-1} \hattilde{\seta}(X) \big\|_1 \Big] \\
   &\le \E_X \Big[ \big\| \T^{-1} \hattilde{\seta}(X) - \T^{-1} \tilde{\seta} (X) \big\|_1 + \big\| \hat{\T}^{-1} \hattilde{\seta}(X) - \T^{-1} \hattilde{\seta}(X) \big\|_1 \Big] \\
   &\le \big\| \T^{-1} \big\|_1 \cdot \E_X \Big[ \big\| \hattilde{\seta}(X) - \tilde{\seta} (X) \big\|_1 \Big] + \big\| \hat{\T}^{-1} - \T^{-1} \big\|_1 \cdot \E_X \Big[ \big\| \hattilde{\seta}(X) \big\|_1 \Big] \\
    &\le \big\| \T^{-1} \big\|_1 \cdot \E_X \Big[ \big\| \hattilde{\seta}(X) - \tilde{\seta} (X) \big\|_1 \Big] + \big\| \hat{\T}^{-1} - \T^{-1} \big\|_1 \, .
\end{align*}
\end{proof}

\begin{lem}[Guarantee for noise-corrected OP2 with estimated $\hat{\T}$]
\label{lem:op2-add}
For $h : \X \to \Delta_n$, let $\hat{\C}^{\tilde{S}}[h]$ be the empirical confusion matrix w.r.t. noisy sample $\tilde{S}$ (computed similarly as $\hat{\C}^{S}[h]$ in Eq. \eqref{eqn:empirical-conf}). Let $\hat{\T}$ be an estimate of $\T$. Then
\begin{align*}
&\big\| \C^D[h] - \hat{\T}^{-1} \hat{\C}^{\tilde{S}}[h] \big\|_{\ve,\infty} \\
&\le n \big\| \T^{-1} \big\|_1 \cdot \big\| \C^{\tilde{D}}[h] - \hat{\C}^{\tilde{S}}[h] \big\|_{\ve,\infty} + \big\| \hat{\T}^{-1} - \T^{-1} \big\|_1 \, .
\end{align*}
\end{lem}

\textbf{Proof of Lemma \ref{lem:op2-add}.}

\begin{proof}
\begin{align*}
&\big\| \C^D[h] - \hat{\T}^{-1} \hat{\C}^{\tilde{S}}[h] \big\|_{\ve,\infty} \\
&= \big\| \T^{-1} \C^{\tilde{D}}[h] - \hat{\T}^{-1} \hat{\C}^{\tilde{S}}[h] \big\|_{\ve,\infty} \\
&= \big\| \T^{-1} \C^{\tilde{D}}[h] - \hat{\T}^{-1} \hat{\C}^{\tilde{S}}[h] + \T^{-1} \hat{\C}^{\tilde{S}}[h] - \T^{-1} \hat{\C}^{\tilde{S}}[h] \big\|_{\ve,\infty} \\
&\le \big\| \T^{-1} \C^{\tilde{D}}[h] - \hat{\T}^{-1} \hat{\C}^{\tilde{S}}[h] + \T^{-1} \hat{\C}^{\tilde{S}}[h] - \T^{-1} \hat{\C}^{\tilde{S}}[h] \big\|_{1} \\
&\le \big\| \T^{-1} \C^{\tilde{D}}[h] - \T^{-1} \hat{\C}^{\tilde{S}}[h] \big\|_{1} + \big\| \T^{-1} \hat{\C}^{\tilde{S}}[h] - \hat{\T}^{-1} \hat{\C}^{\tilde{S}}[h] \big\|_{1} \\
&\le \big\| \T^{-1} \big\|_1 \cdot \big\| \C^{\tilde{D}}[h] - \hat{\C}^{\tilde{S}}[h] \big\|_{1} + \big\| (\T^{-1} - \hat{\T}^{-1}) \hat{\C}^{\tilde{S}}[h] \big\|_{1} \\
&\le n \big\| \T^{-1} \big\|_1 \cdot \big\| \C^{\tilde{D}}[h] - \hat{\C}^{\tilde{S}}[h] \big\|_{\ve,\infty} + \big\| \T^{-1} - \hat{\T}^{-1} \big\|_{1} \cdot \big\| \hat{\C}^{\tilde{S}}[h] \big\|_{1} \\
&\le n \big\| \T^{-1} \big\|_1 \cdot \big\| \C^{\tilde{D}}[h] - \hat{\C}^{\tilde{S}}[h] \big\|_{\ve,\infty} + \big\| \T^{-1} - \hat{\T}^{-1} \big\|_1 \, .
\end{align*}
\end{proof}

\textbf{Proof of Theorem \ref{thm:fw-regret-add}.}

\begin{proof}
Similar as in the proof of Theorem \ref{thm:fw-regret}, chaining Lemma \ref{lem:op1-add} and Lemma \ref{lem:op2-add} with Proposition \ref{thm:fw-regret-standard} establishes the claim.
\end{proof}

\textbf{Proof of Theorem \ref{thm:bs-regret-add}.}

\begin{proof}
Similar as in the proof of Theorem \ref{thm:bs-regret}, chaining Lemma \ref{lem:op1-add} and Lemma \ref{lem:op2-add} with Proposition \ref{thm:bs-regret-standard} establishes the claim.
\end{proof}

\section{Synthetic Data: Additional Details}
\label{appendix:synthetic}

\textbf{Data generating process.}
Specifically, we constructed a 3-class problem over a 2-dimensional instance space $\X = \R^2$ as follows. 
Instances $\x$ were generated according to a fixed Gaussian mixture distribution. The class probability function $\seta:\X\>\Delta_3$ was $\eta_y(\x) = \frac{\exp(\w_y^\top \x + b_y)}{\sum_{y'=1}^3 \exp(\w_{y'}^\top \x + b_{y'})}$ for some fixed weight vectors $\w_1,\w_2,\w_3\in\R^{2}$ and bias terms $b_1, b_2, b_3 \in \R$. 
Given an instance $\x$, a clean label $y$ was drawn randomly according to $\seta(\x)$.
Then $y$ was flipped to a noisy label $\tilde{y}$ according to the probabilities in the $y$-th column of $\T$, where $\T$ is a prescribed column stochastic noise matrix.

\section{Real Data: Additional Details and Experiments}
\label{appendix:real}

For NCFW, NCBS, FW, and BS, the linear model CPE learner was implemented using scikit-learn \citep{scikit-learn}.
The different variants of NCLR, implemented in TensorFlow \citep{tensorflow2015-whitepaper}, also used a linear function class.
In all cases, regularization parameters were chosen by cross-validation.

Table \ref{tab:dataset-info} gives details of the real data sets used in Section \ref{sec:expts}. Table \ref{tab:G-results} and Table \ref{tab:Q-results} give results for G-mean loss and Q-mean loss, respectively.

\begin{table}[t]
\vspace{-6pt}
\begin{center}
\caption{Details of Data Sets Used in Section \ref{sec:expts}}
\label{tab:dataset-info}
\vspace{1pt}
\begin{tabular}{@{}lrrrrr@{}}
\hline
\textbf{{Data set}}		& \textbf{{\# instances}} & \textbf{{\# classes}} 	& \textbf{{\# features}}  \\
\hline
vehicle 	 	& {846} 	& {4}    & {18} \\
pageblocks 	 	& {5,473} 	& {5}    & {10} \\
satimage 	 	& {6,435} 	& {6}    & {36} \\
covtype 	 	& {581,012} 	& {7}    & {14} \\
abalone 	 	& {4,177} 	& {12}    & {8} \\
\hline
\end{tabular}
\end{center}
\vspace{-16pt}
\end{table}

\begin{table*}[t]
\vspace{-6pt}
\begin{center}
\caption{
Comparison with Other Algorithms for G-mean Loss
}
\label{tab:G-results}
\vspace{1pt}
\scalebox{0.9}{
\begin{tabular}{@{}llrrrr@{}}
\hline
\textbf{Data sets} & \textbf{Algorithms} & $\sigma=0.1$ & $\sigma=0.2$ & $\sigma=0.3$ & $\sigma=0.4$ \\
\hline
vehicle & FW & \textbf{0.230 (0.007)} & {0.249 (0.011)} & \textbf{0.287 (0.009)} & {0.349 (0.011)} \\
& NCFW & {0.231 (0.008)} & \textbf{0.247 (0.012)} & {0.289 (0.009)} & \textbf{0.312 (0.012)} \\
& NCLR-Backward & {0.432 (0.017)} & {0.519 (0.013)} & {0.488 (0.034)} & {0.469 (0.025)} \\
& NCLR-Forward & {0.463 (0.030)} & {0.496 (0.015)} & {0.523 (0.012)} & {0.524 (0.025)} \\
& NCLR-Plug-in & {0.477 (0.016)} & {0.503 (0.012)} & {0.486 (0.021)} & {0.502 (0.013)} \\
\hline
pageblocks & FW & {0.331 (0.026)} & {0.225 (0.009)} & {0.552 (0.081)} & {0.563 (0.046)} \\
& NCFW & \textbf{0.236 (0.040)} & \textbf{0.167 (0.010)} & \textbf{0.503 (0.125)} & \textbf{0.382 (0.018)} \\
& NCLR-Backward & {1.000 (0.000)} & {1.000 (0.000)} & {1.000 (0.000)} & {1.000 (0.000)} \\
& NCLR-Forward & {1.000 (0.000)} & {1.000 (0.000)} & {1.000 (0.000)} & {1.000 (0.000)} \\
& NCLR-Plug-in & {1.000 (0.000)} & {1.000 (0.000)} & {1.000 (0.000)} & {0.898 (0.091)} \\
\hline
satimage & FW & {0.181 (0.004)} & {0.214 (0.004)} & {0.225 (0.005)} & {0.310 (0.005)} \\
& NCFW & \textbf{0.180 (0.005)} & \textbf{0.213 (0.004)} & \textbf{0.212 (0.005)} & \textbf{0.272 (0.005)} \\
& NCLR-Backward & {0.368 (0.010)} & {0.400 (0.013)} & {0.428 (0.020)} & {0.684 (0.072)} \\
& NCLR-Forward & {0.362 (0.009)} & {0.351 (0.005)} & {0.385 (0.011)} & {0.524 (0.014)} \\
& NCLR-Plug-in & {0.439 (0.023)} & {0.486 (0.012)} & {0.550 (0.023)} & {0.638 (0.019)} \\
\hline
covtype & FW & {0.570 (0.001)} & {0.577 (0.001)} & {0.685 (0.004)} & {0.690 (0.005)} \\
& NCFW & \textbf{0.515 (0.001)} & \textbf{0.516 (0.001)} & \textbf{0.546 (0.001)} & \textbf{0.612 (0.004)} \\
& NCLR-Backward & {1.000 (0.000)} & {1.000 (0.000)} & {1.000 (0.000)} & {1.000 (0.000)} \\
& NCLR-Forward & {0.908 (0.021)} & {0.875 (0.003)} & {0.869 (0.002)} & {0.847 (0.004)} \\
& NCLR-Plug-in & {1.000 (0.000)} & {1.000 (0.000)} & {1.000 (0.000)} & {1.000 (0.000)} \\
\hline
abalone & FW & {0.779 (0.008)} & {0.775 (0.002)} & {0.786 (0.005)} & \textbf{0.778 (0.005)} \\
& NCFW & \textbf{0.776 (0.007)} & \textbf{0.766 (0.001)} & \textbf{0.781 (0.007)} & {0.784 (0.005)} \\
& NCLR-Backward & {1.000 (0.000)} & {1.000 (0.000)} & {1.000 (0.000)} & {1.000 (0.000)} \\
& NCLR-Forward & {1.000 (0.000)} & {1.000 (0.000)} & {1.000 (0.000)} & {1.000 (0.000)} \\
& NCLR-Plug-in & {1.000 (0.000)} & {1.000 (0.000)} & {1.000 (0.000)} & {1.000 (0.000)} \\
\hline
\end{tabular}
}
\end{center}
\end{table*}

\begin{table*}[t]
\vspace{-6pt}
\begin{center}
\caption{
Comparison with Other Algorithms for Q-mean Loss
}
\label{tab:Q-results}
\vspace{1pt}
\scalebox{0.9}{
\begin{tabular}{@{}llrrrr@{}}
\hline
\textbf{Data sets} & \textbf{Algorithms} & $\sigma=0.1$ & $\sigma=0.2$ & $\sigma=0.3$ & $\sigma=0.4$ \\
\hline
vehicle & FW & \textbf{0.268 (0.008)} & \textbf{0.287 (0.012)} & {0.313 (0.009)} & {0.367 (0.006)} \\
& NCFW & {0.276 (0.005)} & {0.293 (0.009)} & \textbf{0.310 (0.010)} & \textbf{0.334 (0.011)} \\
& NCLR-Backward & {0.449 (0.011)} & {0.518 (0.010)} & {0.495 (0.031)} & {0.480 (0.022)} \\
& NCLR-Forward & {0.473 (0.023)} & {0.501 (0.011)} & {0.527 (0.011)} & {0.527 (0.023)} \\
& NCLR-Plug-in & {0.486 (0.016)} & {0.504 (0.011)} & {0.489 (0.021)} & {0.507 (0.013)} \\
\hline
pageblocks & FW & {0.352 (0.023)} & {0.276 (0.005)} & {0.499 (0.021)} & {0.547 (0.030)} \\
& NCFW & \textbf{0.267 (0.016)} & \textbf{0.201 (0.004)} & \textbf{0.426 (0.032)} & \textbf{0.466 (0.036)} \\
& NCLR-Backward & {0.684 (0.022)} & {0.683 (0.036)} & {0.776 (0.033)} & {0.715 (0.031)} \\
& NCLR-Forward & {0.850 (0.024)} & {0.800 (0.039)} & {0.812 (0.042)} & {0.861 (0.019)} \\
& NCLR-Plug-in & {0.695 (0.039)} & {0.641 (0.045)} & {0.681 (0.026)} & {0.677 (0.055)} \\
\hline
satimage & FW & \textbf{0.197 (0.005)} & \textbf{0.228 (0.006)} & \textbf{0.246 (0.006)} & {0.323 (0.006)} \\
& NCFW & {0.199 (0.006)} & {0.232 (0.005)} & {0.252 (0.007)} & \textbf{0.312 (0.005)} \\
& NCLR-Backward & {0.390 (0.004)} & {0.403 (0.007)} & {0.412 (0.005)} & {0.443 (0.002)} \\
& NCLR-Forward & {0.386 (0.005)} & {0.380 (0.003)} & {0.393 (0.004)} & {0.425 (0.001)} \\
& NCLR-Plug-in & {0.425 (0.004)} & {0.435 (0.003)} & {0.465 (0.003)} & {0.528 (0.006)} \\
\hline
covtype & FW & {0.567 (0.001)} & {0.570 (0.001)} & {0.639 (0.001)} & {0.653 (0.001)} \\
& NCFW & \textbf{0.546 (0.001)} & \textbf{0.544 (0.001)} & \textbf{0.624 (0.000)} & \textbf{0.623 (0.001)} \\
& NCLR-Backward & {0.736 (0.001)} & {0.744 (0.001)} & {0.752 (0.001)} & {0.768 (0.001)} \\
& NCLR-Forward & {0.729 (0.001)} & {0.734 (0.001)} & {0.732 (0.001)} & {0.732 (0.001)} \\
& NCLR-Plug-in & {0.799 (0.000)} & {0.802 (0.000)} & {0.813 (0.000)} & {0.813 (0.000)} \\
\hline
abalone & FW & {0.754 (0.005)} & {0.762 (0.003)} & \textbf{0.763 (0.005)} & \textbf{0.767 (0.003)} \\
& NCFW & \textbf{0.753 (0.005)} & \textbf{0.760 (0.003)} & {0.770 (0.006)} & {0.775 (0.004)} \\
& NCLR-Backward & {0.910 (0.004)} & {0.919 (0.004)} & {0.917 (0.006)} & {0.910 (0.010)} \\
& NCLR-Forward & {0.892 (0.008)} & {0.921 (0.004)} & {0.902 (0.004)} & {0.910 (0.006)} \\
& NCLR-Plug-in & {0.907 (0.006)} & {0.915 (0.007)} & {0.911 (0.005)} & {0.908 (0.006)} \\
\hline
\end{tabular}
}
\end{center}
\end{table*}

\section{Computing Resource}
We ran all experiments on a desktop with one AMD Threadripper 3960X CPU and one Nvidia GeForce RTX 3090 GPU.

\vfill

\end{document}